%% file: egpaper_final.tex
\documentclass[10pt,twocolumn,letterpaper]{article}

\usepackage{iccv}
\usepackage{times}
\usepackage{epsfig}
\usepackage{graphicx}
\usepackage{amsmath}
\usepackage{amssymb}

\usepackage{graphicx}
\usepackage{amsmath}
\usepackage{amssymb}
\usepackage{booktabs}

\usepackage{float} 
\usepackage{multirow}
\usepackage{threeparttable}
\usepackage{makecell}
\usepackage{pifont}
\usepackage[percent]{overpic}
\usepackage{makecell}
\usepackage{bbding}
\usepackage{xcolor}
\usepackage{multirow}
\usepackage{threeparttable}
\usepackage{makecell}
\usepackage{pifont}
\usepackage[percent]{overpic}
\usepackage{makecell}
\usepackage{bbding}
\usepackage{arydshln} 
\usepackage{wrapfig}
\usepackage{amsfonts,amssymb}

\usepackage[accsupp]{axessibility} 

\usepackage[breaklinks=true,bookmarks=false]{hyperref}

\iccvfinalcopy 

\def\iccvPaperID{1214} 

\ificcvfinal\fi

\begin{document}


\title{TORE: Token Reduction for Efficient Human Mesh Recovery with Transformer}

\author{Zhiyang 
Dou$^{1\dag}$ \quad Qingxuan Wu$^{2\dag}$ \quad  Cheng Lin$^{3\ddag}$ \quad Zeyu Cao$^{4\ddag}$ \quad Qiangqiang Wu$^5$ \\[0.3em]
Weilin Wan$^{1}$ \quad Taku Komura$^1$ \quad Wenping Wang$^6$ \\[0.6em]
$^1$The University of Hong Kong \quad $^2$University of Oxford \quad $^3$Tencent Games\\
$^4$University of Cambridge  \quad $^5$City University of Hong Kong \quad $^6$Texas A\&M University}

\maketitle

\let\thefootnote\relax\footnotetext{$\dag, \ddag$ denote equal contributions.}

\ificcvfinal\thispagestyle{empty}\fi
\input{Main/Abstract}
\input{Main/Intro}
\input{Main/Related}
\input{Main/Method}

\input{Main/Results}
\input{Main/Conclusion}

\input{Main/Appendix}

{\small
\bibliographystyle{ieee_fullname}
\bibliography{egbib}
}

\end{document}

%% file: Main/Abstract.tex
\begin{abstract}
In this paper, we introduce a set of simple yet effective TOken REduction (TORE) strategies for Transformer-based Human Mesh Recovery from monocular images. Current SOTA performance is achieved by Transformer-based structures. However, they suffer from high model complexity and computation cost caused by redundant tokens. We propose token reduction strategies based on two important aspects, i.e., the 3D geometry structure and 2D image feature, where we hierarchically recover the mesh geometry with priors from body structure and conduct token clustering to pass fewer but more discriminative image feature tokens to the Transformer. Our method massively reduces the number of tokens involved in high-complexity interactions in the Transformer. This leads to a significantly reduced computational cost while still achieving competitive or even higher accuracy in shape recovery. Extensive experiments across a wide range of benchmarks validate the superior effectiveness of the proposed method. We further demonstrate the generalizability of our method on hand mesh recovery. Visit our project page at \url{https://frank-zy-dou.github.io/projects/Tore/index.html}.
\end{abstract}

%% file: Main/Intro.tex
\vspace{-5mm}
\section{Introduction}
\label{sec:intro}

Human Mesh Recovery (HMR) has been extensively researched in recent years, given its wide real-world applications~\cite{gan2016recognizing,zhai2020two,dong2019towards,sridhar2021class,wang2022towards, pnas_guo, LIU2022129233, ZHANG2023132069}. This task becomes more challenging when the input is a monocular 2D image, due to the large pose and shape variation, large appearance variation, partial observation, and self-occlusion. 

There has been steady progress in 3D human mesh recovery~\cite{kolotouros2019spin, moon2020i2l, li2021hybrik, hmrKanazawa17, sengupta2021hierarchical, lin2021end,lin2021mesh, cho2022FastMETRO, cai2022humman}. Recently, Transformer~\cite{vaswani2017attention} has shown state-of-the-art (SOTA) results on a wide variety of tasks due to its strong capability of capturing long-range dependency for more accurate predictions~\cite{bahdanau2014neural,vaswani2017attention, xia2021soe, Wu_2023_CVPR}.
Using tokens constructed from local features extracted by a convolutional neural network (CNN)~\cite{he2016deep,wang2020hrnet} to query the joint and mesh vertex positions, Transformer-based methods~\cite{lin2021end,lin2021mesh, cho2022FastMETRO} achieved SOTA performance.
\begin{figure}
\centering
\begin{overpic}
[width=\linewidth]{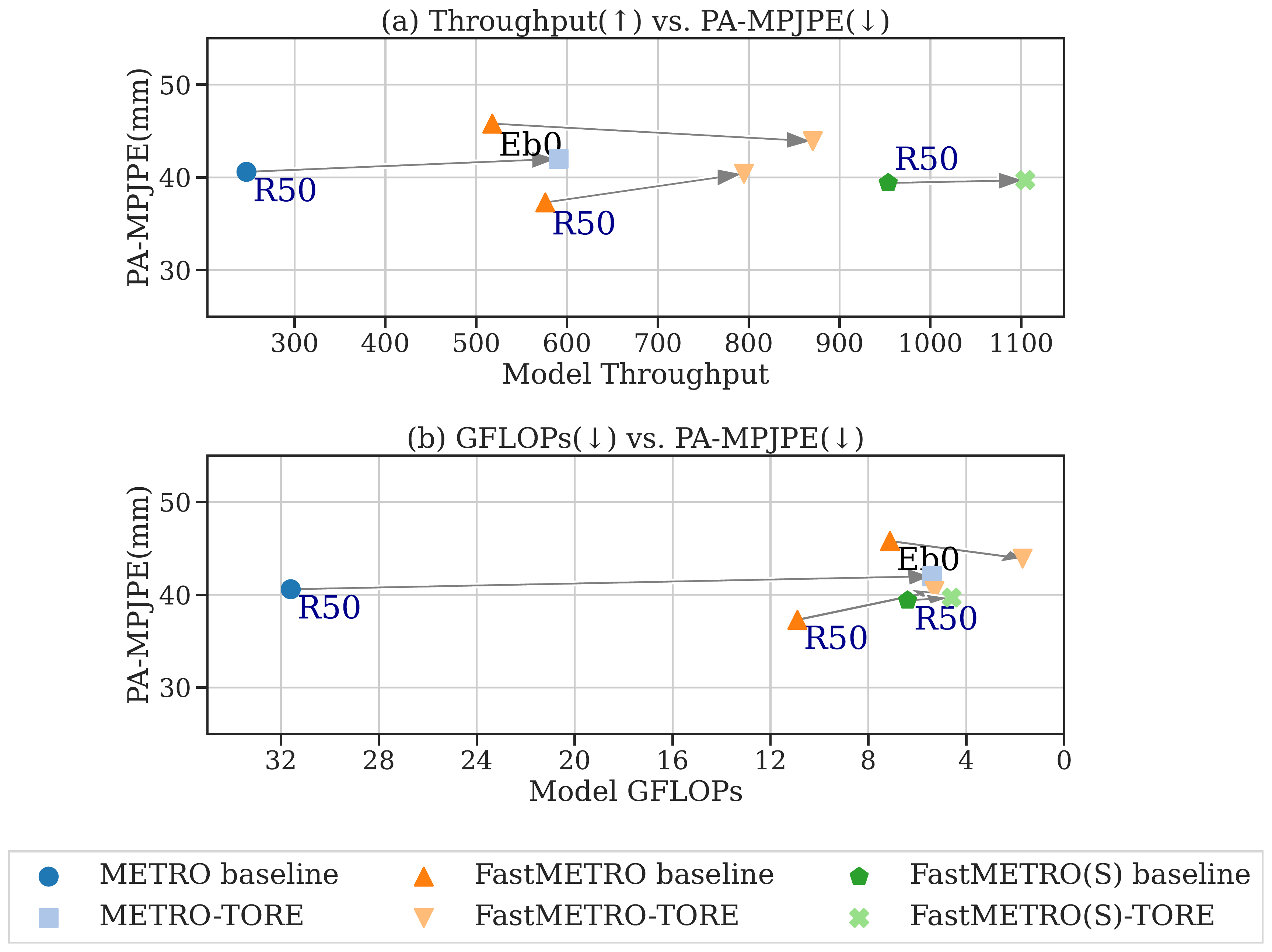} 
\end{overpic}
\vspace{-5mm}
\caption{Throughput v.s. Accuracy and GFLOPs v.s. Accuracy on Human3.6M~\cite{h36m_pami}. Our method dramatically saves GFLOPs and improves throughput while maintaining highly competitive accuracy. The x-axis of the bottom GFLOPs figure is reversed for demonstration. Eb0 and R50 represent EfficientNet-b0~\cite{tan2019efficientnet} and ResNet-50~\cite{he2016deep} backbones, respectively.}
\vspace{-7mm}
\label{fig:teasar}
\end{figure}
 
However, improved performance comes with costs: the increased expressivity of Transformers comes with quadratically increasing computational costs since all pairwise interactions are taken into account~\cite{esser2021taming}. The space and time complexity of a QKV-attention operation is known to be $O(M^2)$, where $M$ is the number of tokens. The token number thus plays a vital role in time efficiency: a large number of tokens inevitably leads to a heavy computation burden. Unfortunately, almost all the existing Transformer-based SOTA methods for HMR~\cite{lin2021end, lin2021mesh, cho2022FastMETRO} are suffering from redundant tokens. This incurs a high model complexity and computational costs, which prevents the current Transformer-based HMR methods from achieving their full potential in real-world applications.

In this paper, we make key observations from two important aspects: the 3D geometry structure and 2D image feature, to reveal the problem of token redundancy. First, to recover the 3D geometry, all existing methods use both mesh vertices and skeleton joints as tokens for the feature interaction between input and body shape. Whereas a body mesh contains numerous vertices, they can be abstracted by a small number of joints on the skeleton. 
For instance, when animating a SMPL~\cite{loper2015smpl} avatar, the skeleton joints, together with a blend shape binding the joints and corresponding mesh vertices of the local body part, are able to describe various body meshes. Therefore, the joints can already be viewed as an underlying structure of a body shape, which intrinsically encodes the human mesh geometry. Second, for image-based input, most existing methods indiscriminately use all the feature patches to capture pose, shape and appearance variance. However, although the human body exhibits large variance, the important features for shape inference are dominantly clustered within the body area in an RGB image. Most features, e.g., image background, are not informative, thus bringing about redundancy. 

Given the aforementioned insights, we argue that the Pareto-front of accuracy and efficiency for Transformer-based HMR could be further improved by reducing the number of tokens ~\cite{tklr2021, ma2022ppt, rao2021dynamicvit}. To this end, we introduce a set of simple yet effective token reduction strategies mainly from two aspects corresponding to our observations. First, for 3D mesh recovery, instead of querying both vertices and joints with input features simultaneously, we consider learning a small set of body tokens at the skeleton level for each body part. To recover corresponding mesh vertices, we use an efficient Neural Shape Regressor~(NSR) to infer the mesh from the body features encoded by these tokens. This query process can also be interpreted as an attention matrix decomposition, by which we effectively leverage the geometric insights encoded at the skeleton level to infer the mesh structure hierarchically. Second, for the input image feature, we introduce a learnable token pruner to prune the tokens of patch-based features extracted by a CNN. We employ a clustering-based strategy to identify discriminative features, which results in two appealing properties: 1) the end-to-end learning of the pruner is unsupervised, avoiding the need for additional data labeling; 2) it learns semantically consistent features across various images, thus further benefiting the geometry reasoning and enhancing the capability of generalizability. These token reduction strategies substantially reduce the number of query tokens involved in the computation without sacrificing the important information. An overview is shown in Figure~\ref{fig:pipeline}.

We conduct extensive experiments across wide benchmarks~\cite{h36m_pami,vonMarcard2018, zimmermann2019freihand}, including both the human body and hand mesh recovery, to validate the proposed method. Compared to SOTA methods, our framework faithfully recovers body meshes with fewer tokens, which considerably reduces memory and computation overhead while maintaining competitive geometric accuracy.

In summary, our contribution is three-fold:
\begin{itemize}
 \item We reveal the issues of token redundancy in the existing Transformer-based methods for HMR.
 \item  We propose effective strategies for token reduction by incorporating the insights from the 3D geometry structure and 2D image feature into the Transformer design.
 \item Our method achieves SOTA performance on various benchmarks with less computation cost. For instance, for the Transformer Encoder structure~\cite{lin2021end} and the Transformer Encoder-Decoder structure~\cite{cho2022FastMETRO} with ResNet-50~\cite{he2016deep} backbone, our method maintains competitive accuracy while saving $82.9\%$, $51.4\%$ GFLOPs and improving $139.1\%$, $38.0\%$ throughput, respectively; see Figure~\ref{fig:teasar} for an overview.  \vspace{-1mm}
 \end{itemize}

%% file: Main/Related.tex
\section{Related Work}
\label{sec:related}
\begin{figure*}
\centering
\begin{overpic}
[width=0.93\linewidth]{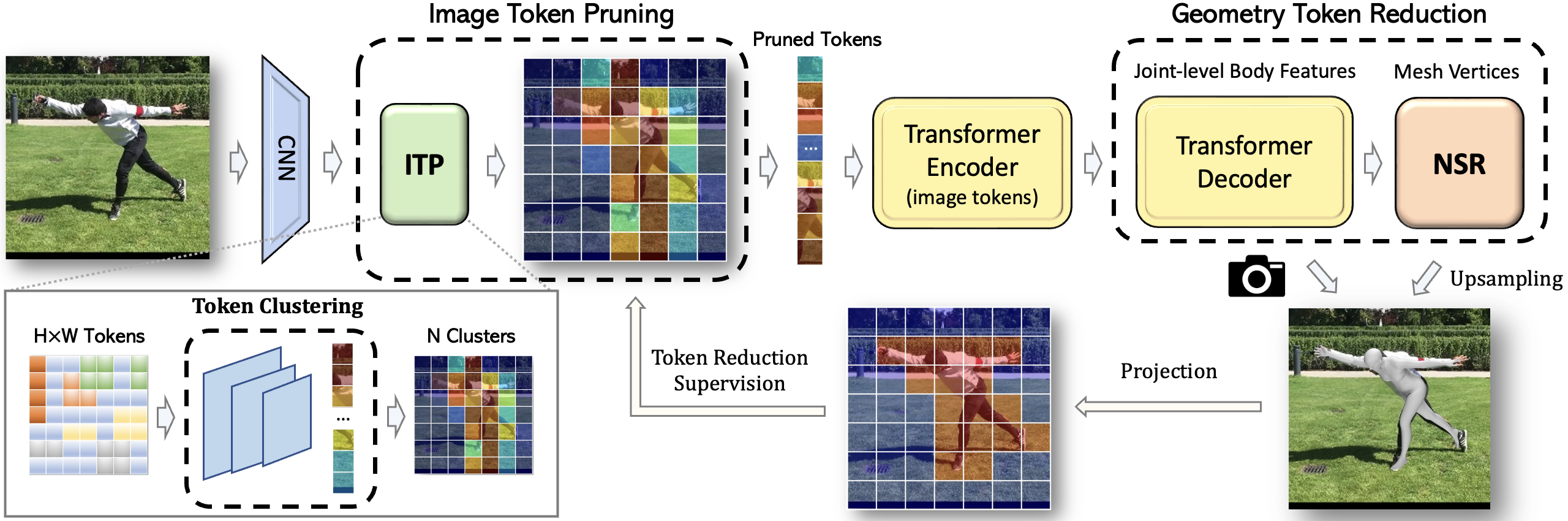} 
\end{overpic}
\caption{Overview of the proposed framework. Our goal is to reduce tokens for Transformer Encoder and Decoder which are critical modules in the whole pipeline. Image Token Pruner~(ITP) and Neural Shape Regressor~(NSR) are two lightweight components.}
\label{fig:pipeline}
\end{figure*}
Human Mesh Recovery (HMR) from monocular images has achieved great progress in the past years~\cite{bogo2016keep,  kolotouros2019spin, moon2020i2l, li2021hybrik, mmhuman3d, lin2021mesh, hmrKanazawa17, sengupta2021hierarchical, cai2022humman, lin2021end, cho2022FastMETRO, zhang2021pymaf,wang2023zolly, rong2021monocular}. Given a monocular RGB image, the goal of HMR is to recover the 3D body shape, typically using a human body model, e.g., SMPL~\cite{loper2015smpl}. We refer readers to~\cite{tian2022recovering} for a comprehensive review. Existing methods fall into two categories: Non-Transformer-based and Transformer-based approaches.\\
\noindent
\textbf{Non-Transformer-based Human Mesh Recovery} To recover the human body shape, one could predict a few parameters, i.e., joint rotation and body shape, to drive the parametric model (parametric approach) or directly regress the vertex positions of the body (non-parametric approach). \\
\noindent 
\textit{Parametric approaches}~SMPLify~\cite{bogo2016keep} estimates human pose and shape by fitting SMPL to the detected 2D key points~\cite{pishchulin2016deepcut} by optimization. Kanazawa et al.~\cite{hmrKanazawa17} adopt adversarial prior knowledge of the 3D body shape into the neural network for HMR. SPIN~\cite{kolotouros2019spin} then combines regression-based and optimization-based methods during the training loop. Intermediate features such as key points~\cite{pavlakos2018learning}, pixel-to-surface correspondences~\cite{xu2019denserac} and texture consistency~\cite{pavlakos2019texturepose}, have been exploited. 
HybrIK~\cite{li2021hybrik} proposes a hybrid inverse kinematics via twist-and-swing decomposition~\cite{baerlocher2001parametrization}. Regressing the body parameters from the image is a highly non-linear mapping~\cite{moon2020i2l}, which thus limits the performance of parametric-based approaches.\\
\textit{Non-parametric approaches} Non-parametric approaches~\cite{moon2020i2l, kolotouros2019convolutional, choi2020pose2mesh} recover spatial coordinates of a body shape directly from the image features. 
I2L-MeshNet ~\cite{moon2020i2l} predicts the
per-pixel likelihood on 1D heatmaps for vertex coordinates for a trade-off between accuracy and computational cost. 
GraphCMR~\cite{kolotouros2019convolutional} and Pose2Mesh~\cite{choi2020pose2mesh} employ graph convolutional neural network (GCNN)~\cite{kipf2016semi} to regress mesh positions by modeling local interactions among vertices from the image~\cite{kolotouros2019convolutional} and 2D pose~\cite{choi2020pose2mesh}, respectively. Although steady progress has been made, SOTA accuracy is produced by Transformer-based methods. \\
\noindent
\textbf{Transformer-based Human Mesh Recovery} Recently, Transformer structures have been successfully adopted in various vision tasks~\cite{dosovitskiy2020image,liu2021swin,carion2020end,panteleris2022pe,tian2022recovering, zeng2022not}.
In HMR, the SOTA performance is achieved by Transformer-based (attention mechanism) approaches~\cite{lin2021end, kocabas2021pare, lin2021mesh, cho2022FastMETRO}.  
Specifically, METRO~\cite{lin2021end} recovers body mesh using a Transformer Encoder~\cite{vaswani2017attention} to model vertex-vertex and vertex-joint interactions conducting dimension reduction from the image features to 3D shape. MeshGraphormer~\cite{lin2021mesh} injects Graph Convolutions~\cite{kipf2016semi} into the Transformer Encoder~\cite{lin2021end} blocks to improve local interactions among vertices. The recent work FastMETRO~\cite{cho2022FastMETRO} employs a Transformer Encoder-Decoder architecture disentangling the image encoding and mesh estimation for model reduction and acceleration. However, all Transformer-based approaches~\cite{lin2021end, lin2021mesh, cho2022FastMETRO} suffer from redundant tokens, causing heavy interaction between body geometry and image features, which is still cumbersome and computationally intensive. In this paper, we design strategies driven by the 3D body geometry and 2D image feature for token reduction for both Transformer Encoder and Transformer Encoder-Decoder structures.\\
\noindent
\textbf{Token Reduction for Transformers} 
Token reduction for Transformer structures has been causing people's attention nowadays~\cite{rao2021dynamicvit,ma2022ppt, tklr2021, yuan2021tokens,grainger2022learning,liang2022not, ma2021transfusion}. In particular, DynamicViT~\cite{rao2021dynamicvit} reduces less informative tokens hierarchically with supervision from a teacher network to save computational costs during inference. TokenLearner~\cite{tklr2021} generates a smaller number of tokens with spatial attention adaptively. Patch-to-Cluster~\cite{grainger2022learning} proposes to cluster over image patches with a token selector. Our method follows this way and adopts a token selection layer for clustering over input tokens, reducing tokens and improving generalization capability in challenging scenarios. Token pruning techniques have also been explored in single-view 2D pose estimation and multi-view 3D pose estimation~\cite{ma2022ppt} by selecting tokens based on the attention value in the Transformer. Differently, in HMR, besides the key points, body geometry, i.e., mesh vertices, also introduces a burden to the Transformer and needs to be carefully considered.

%% file: Main/Method.tex
\section{Methodology}
We employ the popular body parametric model SMPL~\cite{loper2015smpl} to represent a human body. Formally, given a monocular image $\mathcal{I} \in \mathbb{R}^{224\times 224}$, our goal is to recover the joint positions $\mathcal{J} \in \mathbb{R}^{J\times3}$ and mesh vertices $\mathcal{V} \in \mathbb{R}^{V\times3}$ where $J=14, V=6890$ are the numbers of joints and the mesh vertices, respectively. In addition to the human body, we also demonstrate our method on hand mesh recovery with MANO model~\cite{MANO:SIGGRAPHASIA:2017} to show its generalizability.

Existing Transformer-based HMR methods fall into two categories: a single Transformer Encoder structure ~\cite{lin2021end, lin2021mesh} and a Transformer Encoder-Decoder structure~\cite{cho2022FastMETRO}. Accordingly, we propose two strategies to reduce tokens from 3D and 2D levels, namely, Geometry Token Reduction (GTR) and Image Token Pruning (ITP), which can accommodate different Transformer structures to improve their efficiency. 
\subsection{Geometry Token Reduction (GTR)}
As an essential module, all the existing Transformer-based methods for HMR ~\cite{lin2021end, lin2021mesh, cho2022FastMETRO} have to model the interaction between body geometry and image features with an attention mechanism. Therefore, an effective token reduction method to improve the efficiency for inferring mesh geometry can benefit the design of Transformer based methods in a wide scope. 

We observe existing methods invariably use both skeleton joints and mesh vertices as tokens for regressing their spatial coordinates. The numerous tokens inevitably lead to significant computational costs for a Transformer. In fact, as an underlying structure of the human body, the skeleton joints already provide strong priors for perceiving the body geometry. A human body avatar, like SMPL~\cite{loper2015smpl}, can be driven by a small number of joints with the blend shape. This inspires us to regress the human body mesh in a hierarchical manner.

Our key idea is to decompose a heavy Transformer into smaller Transformer modules, each of which involves fewer tokens, avoiding expensive computation for image feature interaction. 
Specifically, our approach only queries the body tokens, whose count is equivalent to the number of joints. Since the number of joints is much less than the number of mesh vertices, we can significantly reduce the token number in complex 2D-3D feature interactions and this leads to a substantial enhancement in efficiency.

To faithfully recover body shape from the body tokens, we introduce a \textit{Neural Shape Regressor (NSR)} module, which is a light Transformer structure, to query the vertex tokens from their interaction with body features. Let $F_J = \{f^j_1, f^j_2,..., f^j_J\}$  
denotes a set of learned body features and $T_\mathcal{V} = \{t^v_1, t^v_2,..., t^v_V\}$ where $t^v_i$ represents a set of vertex tokens which are the query tokens.  
We conduct cross-attention between query vertex tokens and body features to model the interaction between vertices and learned body features. Note that non-adjacent vertices are masked to improve the efficiency ~\cite{cho2022FastMETRO}. 

The feature size within NSR is even smaller compared with the main Transformer. Thanks to the informative geometric feature encoded in the body tokens, despite not incorporating image tokens, NSR still faithfully recovers the surface vertices. Another interesting discovery is that the learned attention scores reflect the correlation between joints and vertices at a body part level, similar to body blending weights (see Sec.~\ref{ablation:VJ_interaction}).  In this way, we can effectively reduce the cost of redundant interaction by decomposing the entire body into encoded constituent parts and efficiently recovering the whole body shape.

\subsection{Image Token Pruning (ITP)}
\label{sec:ITP}

In addition to the geometric insights, for a Transformer Encoder-Decoder structure, image tokens also affect the computation overhead. However, the existing methods adopting this Transformer structure fail to avoid the token redundancy issue~\cite{cho2022FastMETRO} because all available image feature patches are indiscriminately involved in human mesh recovery. Actually, some features, e.g., image background, are not informative, thus introducing redundancy and increasing the computational cost of the Transformer.

To tackle this issue, we introduce an effective token pruning strategy, namely, Image Token Pruning (ITP). Our key insight is that the informative features in an image for inferring 3D geometry are overwhelmingly clustered within the region of the human body. Inspired by recent advances in token pruning~\cite{grainger2022learning, tklr2021, rao2021dynamicvit, ma2022ppt}, we propose 
to aggregate features into a small number of meaningful clusters. 

Let $0< \rho<1$ denote the predefined token pruning ratio. Our goal is to learn a projection that maps feature map $F_\mathcal{I} \in R^{HW \times c}$ extracted from the given image with $HW$ tokens to $Z_\mathcal{I} \in R^{T \times c}$ with $T$ tokens, where $T = \lfloor \rho HW \rfloor$ and $c$ is the feature dimension. The small number of clusters $Z_\mathcal{I}$ are expected to capture the fewer but more discriminative features in $F_\mathcal{I}$. Inspired by \cite{grainger2022learning, tklr2021}, we implement the projection as a learnable CNN module. We first apply $\text{Conv2D}$ to the input feature, where the kernel size, stride and zero padding of $\text{Conv2D}$ are $3 \times 3$, $1$ and $1$, reducing the feature dimension from $c$ to $c' = c/4$ with GELU~\cite{hendrycks2016gaussian} activation function applied. We further map the dimension $c$ to $N$ by $\text{MLP}$ and $\text{Softmax}$, producing $M \in R^{HW \times N}$.  Finally, the clustered token set is given by $Z_\mathcal{I}=M^T \cdot F_\mathcal{I}$. The mapping matrix $M^T$, in essence, produces a clustering over origin tokens. Each element $m_{ij} \in M^T$ depicts the contribution of $j$ token to $i$-th cluster. Note that \text{LayerNorm} is employed after the clustering.

\noindent
\textbf{Token Reduction Supervision} 
In order to encourage ITP to pay more attention to those discriminative feature regions, i.e., body parts. during token reduction, we introduce Token Reduction Supervision. Given the weak-perspective camera estimated the neural network, we compute the 2D projection of the ground truth 3D vertices $ \hat{\mathcal{V}}$ as $s\Pi(\hat{\mathcal{V}})+ t$  where 
$\Pi$ represents an orthographic projection, $s$ and $t$ are scale and translation estimated by the network. The projected results are downsampled to a discrete $H \times W$ grid corresponding to each token followed by a binary indicator function 
\begin{equation}
    F_d(x)=\left\{
\begin{aligned}
1 &, \text{if the cell contains a projected point.}  \\
0 &, \text{if the cell does not contain a projected point.}  \\
\end{aligned}
\right.
\end{equation}
Then the supervision for token pruning is given by
\begin{equation}
 L_{P} = - \frac{1}{NHW}\sum^N_{i}(F_{d}(s\Pi(\hat{\mathcal{V}})+ t) \cdot M_{\left[:,i\right]}).
\label{eq:projection}
\end{equation}

With this supervision, the weights of background tokens in $M[:,i]$ for each $i$-th cluster will be penalized during training, which thus encourages the ITP to learn the discriminative features.

The proposed ITP has several noticeable advantages. First, in contrast to other methods relying on explicit supervision, e.g., using a pretrained teacher network to facilitate the pruning~\cite{rao2021dynamicvit,yuan2021tokens, kong2022spvit}, our token pruner is trained in an unsupervised manner; Note that no ground truth masking is required for Token Reduction Supervision. Second, unlike previous clustering-based pruning methods~\cite{grainger2022learning} that use pre-defined class labels, ITP aims to adaptively identify discriminative features within the body region. Therefore, ITP is not limited to fixed semantic labels but is able to induce higher-level semantics according to the learning target.

Finally, we find the clustered informative features learned by ITP improve the model generalizability, especially for challenging datasets, e.g., in-the-wild dataset 3DPW~\cite{vonMarcard2018} as shown in Table~\ref{tab:enc_dec_GTR_ETP_accuracy}. We elaborate on these properties in Sec.~\ref{sec:fastmetro_GTR_ITP} and Sec.~\ref{ablation:prune_vis} with extensive experiments. 

\subsection{Loss Functions}
\label{sec:loss}
We supervise the network using L1 distance between predicted mesh vertices at three sampling levels: $L_{\mathcal{V}3D} = ||\mathcal{V}_{3D}^l-\hat{\mathcal{V}}^l||_1+||\mathcal{V}_{3D}^m- \hat{ \mathcal{V}}^m||_1 + 
||\mathcal{V}_{3D}^h- \hat{\mathcal{V}}_h||_1,$
where $*^l,*^m, *^h$ denote low, middle and high body mesh resolution with vertex numbers to be $431$, $1723$ and $6890$ for an SMPL body. The ground-truth 3D joints are used for supervising predicted 3D joints and the ones regressed from the vertices $\mathcal{V}^h_{3D}$ with a SMPL regression matrix $\mathcal{M}$:
$L_{\mathcal{J}3D} = ||\mathcal{J}_{3D} - \hat {\mathcal{J}}_{3D}||_1, L^{R}_{\mathcal{J}3D} = ||\mathcal{M}(\mathcal{V}) - \hat{\mathcal{J}_{3D}} ||_1.$ A 2D projection loss $L_{J2D}$ is used during the network training, where we employ a weak perspective camera model to project 3D joints to 2D for supervision: $L^{R}_{\mathcal{J}2D} = ||(s\Pi(\mathcal{M}(\mathcal{V}))+ t) - \hat{\mathcal{J}_{2D}}||_1,$. $s,t$ are scale and translation estimated by the network. Overall, together with the Token Pruning Supervision $L_{P}$, the total loss is:
\begin{equation*}
\begin{split}
	L &= \alpha \left[\lambda_{\mathcal{J}3D}(L_{\mathcal{J}3D}^R+L_{\mathcal{J}3D}) +  \lambda_{\mathcal{V}3D}(L_{\mathcal{V}3D}) + \lambda_{P}L_{P}\right]  \\
	&+ \beta \lambda_{\mathcal{J}2D}L^R_{\mathcal{J}2D},
\end{split}
\end{equation*} where $\alpha$ and $\beta$ indicate the availability of the supervision. We set $ \lambda_{P}, \lambda_{J2D}, \lambda_{V3D}, \lambda_{J3D}$ to be $1, 100, 100,1000$.

%% file: Main/Results.tex
\section{Experimental Results}
\subsection{Datasets and Metrics}
We evaluate our model in two scenarios: human body mesh recovery and hand mesh recovery. We adopt commonly-used metrics~\cite{kanazawa2018end, kolotouros2019learning,  lin2021end, lin2021mesh,cho2022FastMETRO}:  Mean Per-Joint Position Error (MPJPE), MPJPE after further alignment, i.e., Procrustes Analysis (PAMPJPE) and Mean Per-Vertex Error (MPVE). For human body mesh recovery, our network is trained with Human3.6M~\cite{h36m_pami}, MuCo-3DHP\cite{mehta2018single}, UP-3D~\cite{lassner2017unite}, COCO~\cite{lin2014microsoft}, MPII~\cite{andriluka20142d}. Following previous works~\cite{lin2021end, lin2021mesh, cho2022FastMETRO}, we use the pseudo mesh data in Human3.6M~\cite{h36m_pami} for training, splitting subjects S1, S5, S6, S7, S8 for training and S9, S11 for testing. 
We also report our performance on 3DPW~\cite{vonMarcard2018}, a more challenging in-the-wild dataset. To further evaluate the generalization ability of the proposed method, we test our method on hand mesh recovery on FreiHAND~\cite{zimmermann2019freihand} dataset. For analyzing the efficiency of our method, we report GFLOPs and throughput (image per second), strictly following ~\cite{rao2021dynamicvit, meng2022adavit, tklr2021, ma2022ppt, liang2022not}.

\subsection{Implementation Details}
\label{sec:implementation}

\noindent\textbf{Network Training} We implement the network using PyTorch. For the Transformer Encoder-Decoder structure, we set the learning rate to be~$1 \times 10^{-4}$. We use the AdamW optimizer~\cite{loshchilov2018decoupled} and train for $60$ epochs, with a batch size of $16$ per GPU on $4$ Nvidia A100 GPUs. When comparing with Transformer Encoder structure METRO~\cite{lin2021end} (see Table~\ref{tab:enc_GTR}), we follow the setting of METRO where we train the models with a batch size of $30$ per GPU on $8$ Nvidia A100 GPUs in total. We adopt Adam optimizer~\cite{kingma2015adam} and train the models for $200$ epochs. See more details in Appendix~A.

\noindent\textbf{Performance Evaluation} To analyze the performance on a consumer-level GPU device, we measure the throughput on an NVIDIA RTX 3090 GPU with 24G VRAM. When comparing with encoder-decoder Transformer structure FastMETRO~\cite{cho2022FastMETRO}, we set the batch size to  $32$ following PPT~\cite{ma2022ppt}.
The comparison with Transformer Encoder structure METRO~\cite{lin2021end} uses $16$ as the batch size. Note that the batch size is limited by the size of the original METRO~\cite{lin2021end} model. To factor out the influence of the batch size, we provide a more detailed performance report in Appendix~B.

\subsection{Performance on Human Mesh Recovery}
\label{sec:hmr_results}

Currently, there only exist two types of representative Transformer structures for HMR: Encoder-only structure (METRO~\cite{lin2021end}) and Encoder-Decoder structure
(FastMETRO~\cite{cho2022FastMETRO}). To demonstrate the effectiveness of our method to both Transformer structures, we first conduct experiments by adding GTR to two structures: METRO and FastMETRO. Since ITP is designed for the encoder-decoder structure, we further report the results on FastMETRO with both ITP and GTR in Sec.~\ref{sec:fastmetro_GTR_ITP}.
The overall comparison is shown in Table~\ref{tab:body_mesh_all}.

\subsubsection{Transformer Encoder Structure}
\label{sec:metro_GTR}

\begin{table}
\begin{center}
 \caption{Comparison with the Transformer Encoder structure METRO~(M)~\cite{lin2021end} on Human3.6M~\cite{h36m_pami}. We test with ResNet-50 (R50)~\cite{he2016deep} and HRNet-W64 (H64)~\cite{wang2020hrnet} as backbones. GFLOPs$^\text{T}$ is GFLOPs of the transformer.}
 \label{tab:enc_GTR}
\resizebox{0.485\textwidth}{!}{
\setlength{\tabcolsep}{0.5mm}{
\begin{tabular}{l|lll|l}
\hline
{Method} &  {GFLOPs~ $\downarrow$} & {GFLOPs$^\text{T}$~$\downarrow$} &  {Throughput~(im/s)~ $\uparrow$} &{PAMPJPE~$\downarrow$}\\
\hline
M-H64~\cite{lin2021end}& $56.5$ & $27.5$ &  $141.0$ & $\textbf{36.7}$ \\
M-H64+GTR & $\textbf{30.2~(-46.5\%)}$ & $\textbf{0.8 (-97.1\%)}$ 
& $\textbf{210.1~(+49.0\%)}$  & $37.1~(+1.1\%)$ \\
\hdashline
M-R50~\cite{lin2021end} & $31.6$ &  $27.5$ &  $247.0$ & $\textbf{40.6}$ \\
M-R50+GTR  & $\textbf{5.4~~~(-82.9\%)}$ & $\textbf{0.8~(-97.1\%)}$ 
& $\textbf{590.6~(+139.1\%)}$  & $42.0~(+3.4\%)$ \\ 
 \hline
\end{tabular} 
  }
}
\end{center}
\end{table}
 
\begin{figure}
\centering
\begin{overpic}[width=\linewidth]{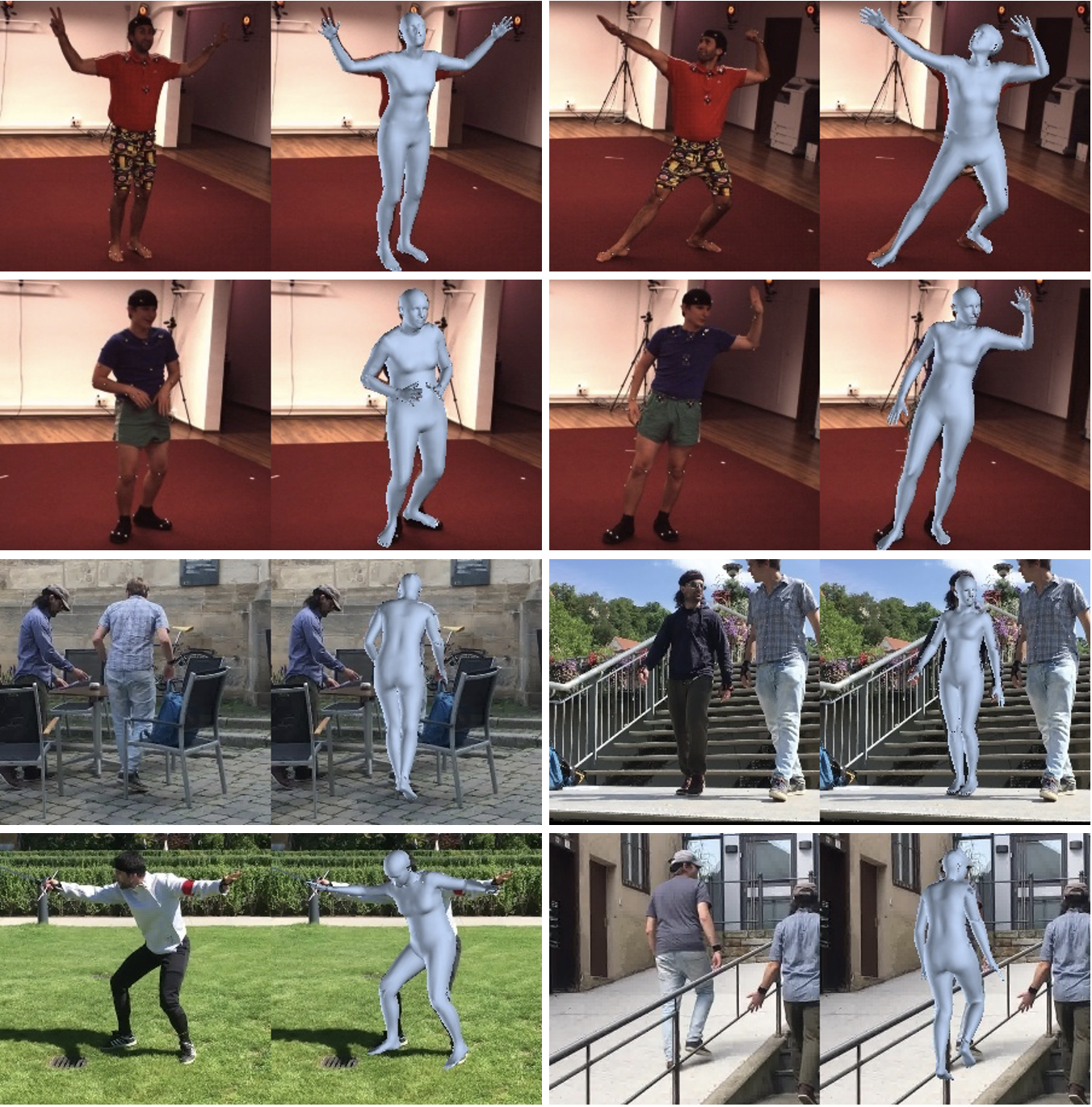}
\end{overpic}
\caption{Qualitative results of GTR equipped Encoder-Decoder structure~\cite{lin2021end} (H64) on Human3.6M~\cite{h36m_pami} and 3DPW~\cite{vonMarcard2018}.}
    \label{fig:body_mesh_metro_dtp}
\end{figure}

As shown in Table~\ref{tab:enc_GTR}, GTR effectively reduces the computation costs while still producing competitive accuracy results. For the encoder-based Transformer model METRO~\cite{lin2021end} with HRNet-W64~\cite{wang2020hrnet} as a backbone, applying GTR saves the GFLOPs of the whole model for $46.5\%$ and the Transformer part for $97.1\%$, with throughput improved by $49\%$. For the ResNet-50~\cite{he2016deep} backbone, GTR helps save the GFLOPs of the whole model for $82.9\%$ and the Transformer part for $97.1\%$, with throughput improved by $139.1\%$. These results validate the effectiveness of the proposed GTR on the Transformer Encoder structure. Qualitative results of METRO-H64+GTR can be found in Figure~\ref{fig:body_mesh_metro_dtp}, in which the model produces high-quality results in human mesh recovery over various input monocular images.

\subsubsection{Transformer Encoder-Decoder Structure} 
\label{sec:fastmetro_GTR_ITP}
We experiment on the Transformer Encoder-Decoder structure FastMETRO~\cite{cho2022FastMETRO} with its two variants: models with $1$ and $3$ Encoder-Decoder layers. For $1$-layer FastMETRO, we denote it as FastMETRO(S).
\begin{table}
\begin{center}
 \caption{Comparison with the Transformer Encoder-Decoder structure FastMETRO~(FM)~\cite{cho2022FastMETRO} on Human3.6M~\cite{h36m_pami}. We test with EfficientNet-b0~(Eb0)~\cite{tan2019efficientnet}, ResNet-50 (R50)~\cite{he2016deep} and HRNet-W64 (H64)~\cite{wang2020hrnet} as backbones. GFLOPs$^\text{T}$ stand for GFLOPs for the transformer.}
 \label{tab:enc_dec_GTR}
\resizebox{0.485\textwidth}{!}{
\setlength{\tabcolsep}{0.5mm}{
\begin{tabular}{l|lll|l}
\hline
{Method} &  {GFLOPs~ $\downarrow$} &
{GFLOPs$^\text{T}$~ $\downarrow$} &  {Throughput~(im/s)~ $\uparrow$} &{PAMPJPE~$\downarrow$}\\
\hline
FM-H64~\cite{cho2022FastMETRO} & $35.7$ & $6.6$  & $221.5$  & $\textbf{33.7}$\\ 
FM-H64+GTR & $\textbf{30.2~(-15.4\%)}$ & $\textbf{0.7~(-89.4\%)}$ 
& $\textbf{249.2~~~(+12.5\%)}$  & $34.8~(+3.2\%)$ \\
\hdashline
FM-R50~\cite{cho2022FastMETRO}  & $10.9$ & $6.6$  & $576.0$  & $\textbf{37.3}$\\ 
FM-R50+GTR  & $\textbf{5.4~~~(-50.5\%)}$ & $\textbf{0.7~(-89.4\%)}$  & $\textbf{805.3~~~(+39.8\%)}$  & $38.6~(+3.4\%)$ \\
\hdashline
FM(S)-R50~\cite{cho2022FastMETRO} & $6.4$ & $2.2$ & $953.5$  & $39.4$\\ 
FM(S)-R50+GTR & $\textbf{4.6~~~(-28.1\%)}$ & $\textbf{0.3~(-86.4\%)}$ & $\textbf{1128.9~(+18.4\%)}$  & $\textbf{38.6~(-2.0\%)}$\\ 
\hdashline
FM-Eb0 &  $7.1$ & $6.6$ & $517.6$  & $45.8$\\
FM-Eb0+GTR  &  $\textbf{1.7~~~(-76.1\%)}$ & $\textbf{0.7~(-89.4\%)}$ & $$\textbf{870.5~~~(+68.2\%)}$$  & $\textbf{44.2~(-3.5\%)}$\\
 \hline
\end{tabular} 
  }
  }
\end{center}
 \end{table}
In Table~\ref{tab:enc_dec_GTR}, GTR equipped FastMETRO~\cite{lin2021end} produces competitive accuracy while reducing $15.4\%$ in GFLOPs for the whole model, $89.4\%$ in GFLOPs for the Transformer part and improving $12.5\%$ in throughput with an HRNet-W64~\cite{wang2020hrnet} CNN backbone. When testing with ResNet-50~\cite{he2016deep}, GTR helps save $50.5\%$ in GFLOPs for the whole model, $89.4\%$  in GFLOPs for the Transformer part and improves $39.8\%$ in throughput.  
Notably, our FastMETRO(S)-R50+GTR and FastMETRO(S)-Eb0+GTR yield $38.8, 44.2$ mm PAMPJPE surpassing the corresponding baselines while greatly saving the computational costs.

We then present the performance of the Transformer Encoder-Decoder structure with GTR and ITP using ResNet-50~\cite{he2016deep} and EfficientNet-b0~\cite{tan2019efficientnet} CNN backbones. 

\begin{table}[H]
\small
\begin{center}
 \caption{Influence of ITP for monocular 3D human mesh recovery on 3DPW~\cite{vonMarcard2018}. } \label{tab:enc_dec_GTR_ETP_accuracy}
\resizebox{0.485\textwidth}{!}{
\setlength{\tabcolsep}{3.5mm}{
\begin{tabular}{l|ccc}
\hline
{Method}  &{MPVE} &{MPJPE} &{PAMPJPE}\\
\hline
FastMETRO-H64+GTR & $91.3$ & $75.4$ &	$46.7$ \\
\makecell[c]{FastMETRO-H64+GTR+ITP@$20\%$} & $\textbf{88.2}$&	$\textbf{72.3}$	&$\textbf{44.4}$\\
 \hline
\end{tabular} 
  }
  }
\end{center}
 \end{table}
As shown in Table~\ref{tab:enc_dec_GTR_ETP}, Tore (GTR+ITP) is effective for the Encoder-Decoder Transformer structure. Specifically, when pruning with $20\%, 50\%$ in ITP, the models save the GLOPs of the Transformer structure by $14.3\%, 14.3\%$ and $28.5\%, 42.9\%$ while receiving competitive accuracy or even higher accuracy i.e., FM-Eb0+GTR+ITP@20\% improves PAMPJPE compared with the baseline.

\begin{table*}
\small
\begin{center}
 \caption{Statistics of all proposed components (GTR + ITP) for Encoder-Decoder structure FastMETRO~(FM)~\cite{cho2022FastMETRO} on Human3.6M~\cite{h36m_pami}. The backbones are EfficientNet-b0~(Eb0)~\cite{tan2019efficientnet} and ResNet-50 (R50)~\cite{he2016deep}. GFLOPs$^\text{T}$ stands for GFLOPs of the Transformer.}
 \label{tab:enc_dec_GTR_ETP}
\resizebox{1.0\textwidth}{!}{
\setlength{\tabcolsep}{5.5mm}{
\begin{tabular}{l|c|lll|l}
\hline
{Method + Pruning Rate} & {\#Tokens} &  {GFLOPs~ $\downarrow$} & {GFLOPs$^\text{T}$~ $\downarrow$} &  {Throughput~(im/s)~$\uparrow$} &{PAMPJPE~$\downarrow$}\\
\hline
FM-R50+GTR & $49$  & $5.4$ & $0.7$  & $805.3$  & $\mathbf{38.6}$ \\
FM-R50+GTR+ITP@$20\%$& $39$ & $\textbf{5.3~(-1.8\%)}$ & $\textbf{0.6 (-14.3\%)}$ 
& ${794.7~(-1.3\%)}$  & $40.5~(+4.9\%)$ \\ 
FM-R50+GTR+ITP@$50\%$ & $24$ & $\textbf{5.1~(-5.5\%)}$ & $\textbf{0.5~(-28.5\%)}$ 
& ${{806.1~(+0.1\%)}}$  & $40.7~(+5.4\%)$ \\  
\hdashline
FM-Eb0+GTR & $49$ & $1.7$ & $0.7$ 
& $870.5$  & $44.2$ \\ 

FM-Eb0+GTR+ITP@$20\%$ & $39$ & $\textbf{1.6~(-5.9\%)}$ & $\textbf{0.6~(-14.3\%)}$ 
& ${876.3~(+0.7\%)}$  & $\mathbf{43.9~(-0.6\%)}$ \\ 

FM-Eb0+GTR+ITP@$50\%$ & $24$ & $\textbf{1.4~(-17.6\%)}$ & $\textbf{0.4~(-42.9\%)}$ 
& $870.4~(-0.1\%)$   & $\mathbf{43.9~(-0.6\%)}$ \\ 
 \hline
\end{tabular} 
 }
 }
\end{center}
 \end{table*}

\begin{table}[H]
\small
\begin{center}
 \caption{Comparison with the SOTA methods for monocular 3D human mesh recovery on 3DPW~\cite{vonMarcard2018} and Human3.6M~\cite{h36m_pami}. }
 \label{tab:body_mesh_all}
\resizebox{0.49\textwidth}{!}{
\setlength{\tabcolsep}{0.3mm}{
\begin{tabular}{l|ccc|cc}
\hline
\multirow{2}{*}{Method} & \multicolumn{3}{c|}{3DPW} & \multicolumn{2}{c}{Human3.6M}\\
  & MPVE & MPJPE & PAMPJPE  & MPJPE & PAMPJPE \\
  \hline
HMR~\cite{kanazawa2018end} & – & $130.0$ & $76.7$ & $88.0$ & $56.8$ \\
GraphCMR~\cite{kolotouros2019convolutional} & – &– &$70.2$& – &$50.1$\\
SPIN~\cite{kolotouros2019spin} & $116.4$ & $96.9$ & $59.2$ &$62.5$ & $41.1$\\
I2LMeshNet~\cite{moon2020i2l} & – & $93.2$ & $57.7$ &$55.7$& $41.1$\\
PyMAF~\cite{pymaf2021}& $110.1$ & $92.8$& $58.9$& $57.7$ & $40.5$\\
ROMP–R50~\cite{sun2021monocular} & $105.6$& $89.3$ & $53.5$ &– &–\\
PARE–R50~\cite{kocabas2021pare} & $99.7$ &$82.9$ &$52.3$& – &–\\
DSR–R50~\cite{dwivedi2021learning} & $99.5$ & $85.7$ &  $51.7$ & $60.9$ & $40.3$\\
METRO–R50~\cite{lin2021end} & – & – & – & $56.5$ & $40.6$\\
METRO–H64~\cite{lin2021end} & $88.2$ & $77.1$ & $47.9$ & $54.0$ & $36.7$\\
METRO–H64+GTR & $87.9$ & $75.5$ & $46.6$ & $57.6$ & $37.1$\\
FastMETRO-R50~\cite{cho2022FastMETRO}& $90.6$ & $77.9$ & $48.3$ & $53.9$ & $37.3$ \\
FastMETRO-R50+GTR+ITP@$20\%$ & $ 99.2$ & $ 82.4$ & $ 52.3$ & $ 59.8$ & $40.5$ \\
FastMETRO-H64~\cite{cho2022FastMETRO}& $84.1$ & $73.5$ & $44.6$ & $52.2$ & $33.7$ \\
FastMETRO-H64+GTR+ITP@$20\%$ &   $88.2$&	$72.3$	&$44.4$ & $ 59.6$ & $36.4$ \\
FastMETRO-Eb0~\cite{cho2022FastMETRO}& $112.5$ &$93.8$& $60.2$ & $69.2$ & $45.8$ \\
FastMETRO-Eb0+GTR+ITP@$20\%$ &	$112.5$ & $93.7$ & $60.1$ & $63.2$& $43.9$ \\
 \hline
\end{tabular} 
  }
  }
\end{center}
 \end{table}
 
\begin{figure*}
\centering 
\begin{overpic}[width=\linewidth]{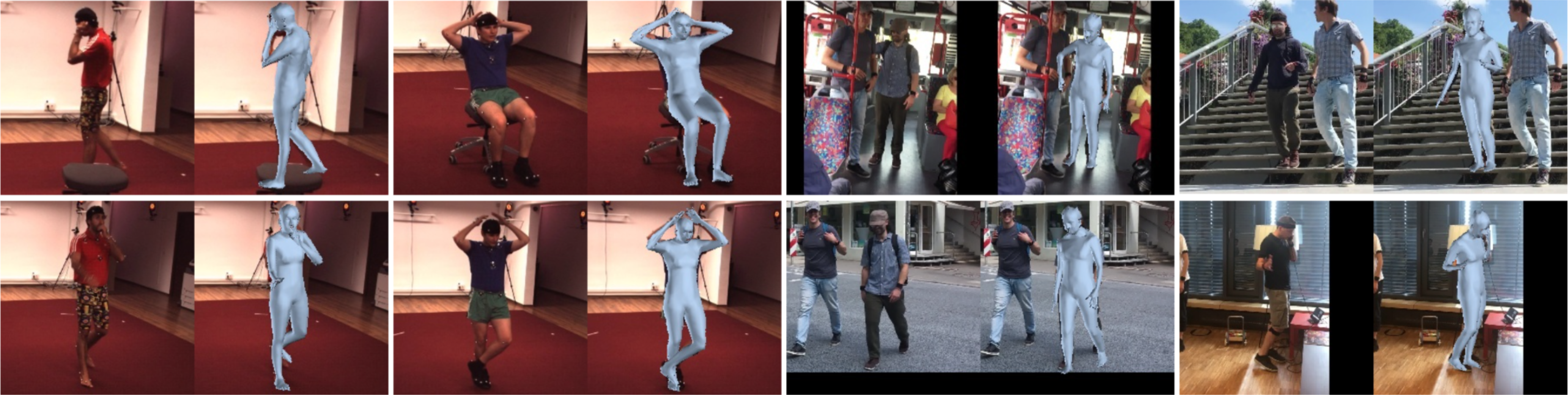}
\end{overpic}
\caption{Qualitative results of FastMETRO+GTP+ITP@$20\%$ on Human3.6M~\cite{h36m_pami} and 3DPW~\cite{vonMarcard2018}.}
\label{fig:full_body_mesh_result}
\end{figure*}

\begin{figure*}
\centering
\begin{overpic}[width=\linewidth]{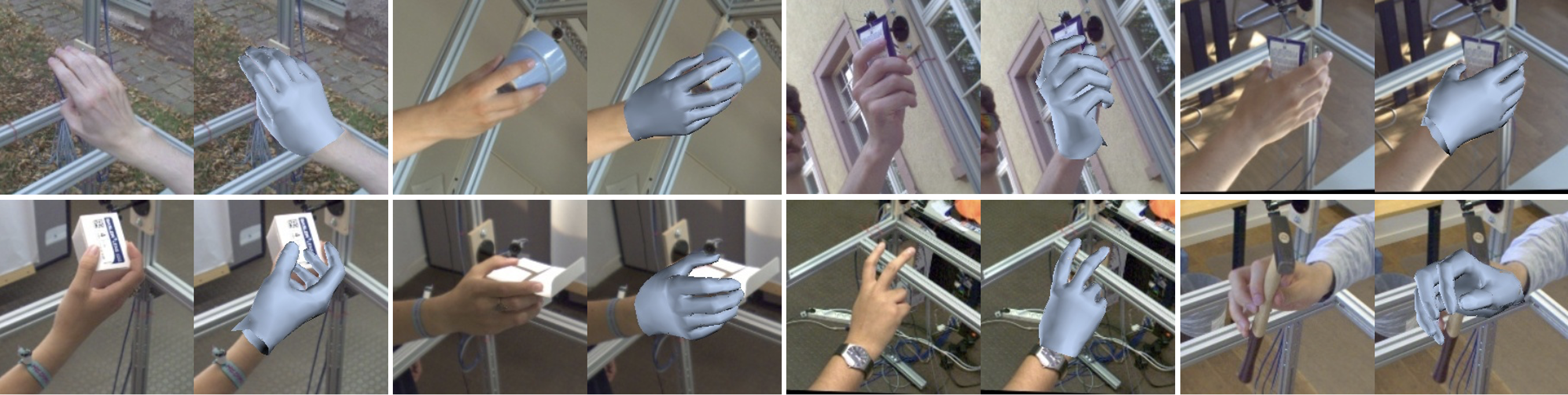}
\end{overpic}
\caption{Qualitative results on FreiHAND~\cite{zimmermann2019freihand} by FastMETRO+H64+GTR+ITP@$20\%$ model.}
\label{fig:hand_mesh_main_result}
\end{figure*}

In addition, we find that ITP helps improve the accuracy on the challenging in-the-wild 3DPW~\cite{vonMarcard2018} dataset, shown in Table~\ref{tab:enc_dec_GTR_ETP_accuracy}. Specifically, when further equipped with ITP, the model performance in MPVE, MPJPE and PAMPJPE are improved by $3.1$mm, $3.1$mm and $2.3$ mm, respectively. These results indicate that the ITP module learns more discriminative features and thus enhances the capability of generalization, allowing methods with ITP to achieve better performance in more challenging in-the-wild scenarios. 

The strong representation capability of ITP can also be seen in the clustering process, which promotes competitive accuracy with fewer tokens while producing semantically meaningful clustering results as demonstrated in Sec.~\ref{ablation:prune_vis}.

We conduct comparisons with existing SOTA approaches (both Transformer and Non-Transformer based) on monocular human mesh recovery. The results are summarized in Table~\ref{tab:body_mesh_all}. It is notable that, with effective token reduction techniques, our method produces competitive or higher accuracy on Human3.6M~\cite{h36m_pami} and 3DPW~\cite{vonMarcard2018} datasets. For instance, on 3DPW, our METRO-H64+GTR achieves $87.9$mm MPVE, $75.5$mm MPJPE and $46.6$mm PAMPJPE surpassing METRO–H64. In general, token reduction results in information loss, which leads to a slight drop in accuracy. For MPVE, using an HRNet-W64 backbone, GTR typically causes a larger error increase of 2.6 mm (from $84.1$ to $86.7$), while ITP is 1.5 mm (from $86.7$ to $88.2$). This suggests GTR sacrifices more accuracy for efficiency, while ITP has less impact. Qualitative results of FastMETRO+GTR+ITP@$20\%$ are visualized in Figure~\ref{fig:full_body_mesh_result}, where our method produces accurate and robust human mesh recovery from monocular images. In the inset, we also provide a qualitative comparison of FastMETRO+GTR+ITP@$20\%$~(w/ GTR+ITP) and the baseline model FastMETRO~(w/o GTR+ITP), \begin{wrapfigure}{r}{2.5cm}
  \hspace*{-4.8mm}
  \centerline{
  \includegraphics[width=30mm]{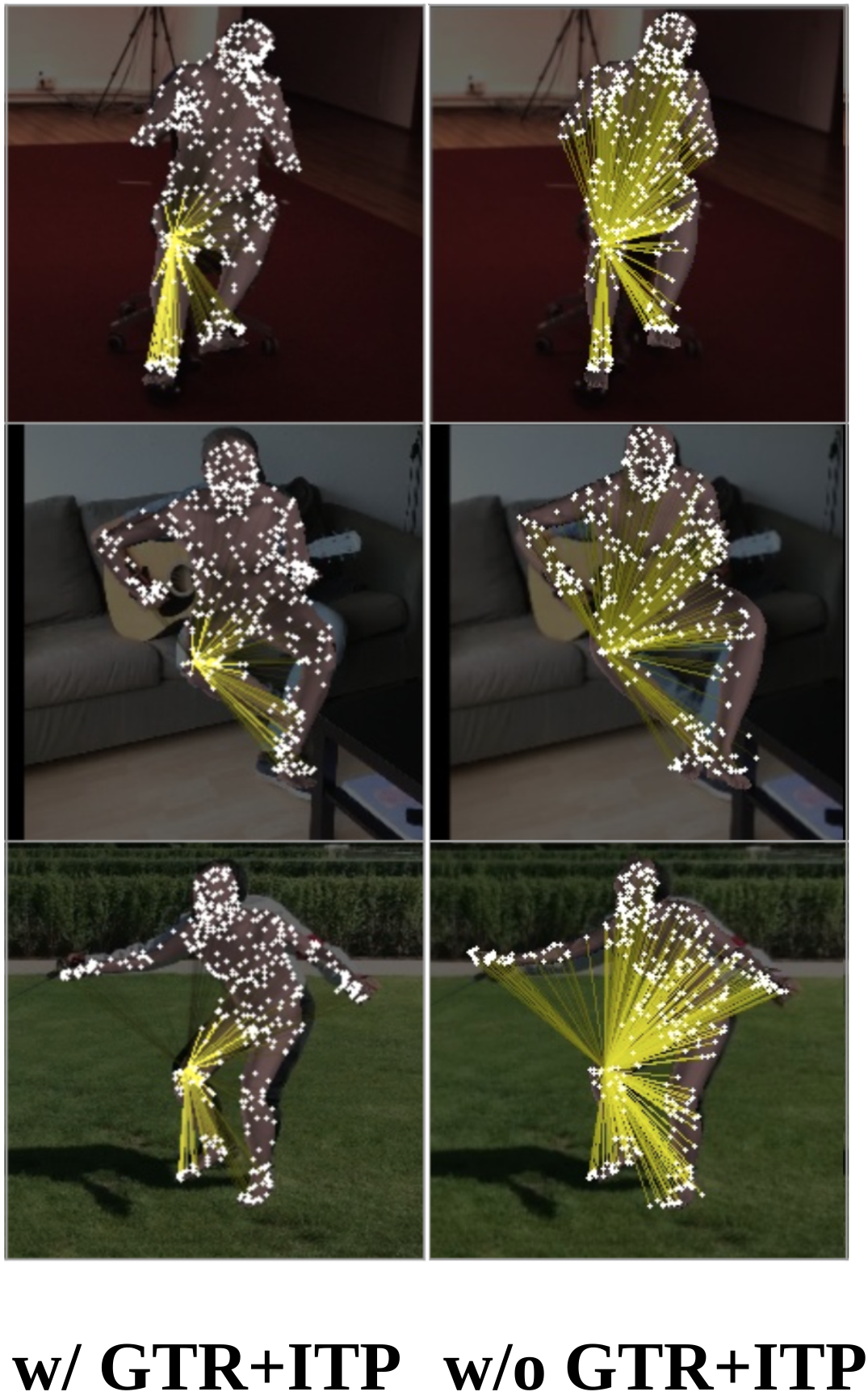}}
  \vspace*{-4mm}
\end{wrapfigure} where the joint-vertex attention is similar to the blending weights in SMPL, which properly captures the shape structure. However, the model w/o GTR+ITP redundantly correlates local joints with distant vertices, leading to additional interaction costs.
In summary, extensive experiments validate the effectiveness of proposed strategies for token reduction across different Transformer structures (Encoder-based METRO~\cite{lin2021end} and Encoder-Decoder-based FastMETRO~\cite{cho2022FastMETRO}) and different Transformer model sizes (FastMETRO and FastMETRO(S)).

\subsection{ITP v.s. TokenLearner~\cite{tklr2021}}
\begin{table}[!h]
\small
 \caption{Comparison with TokenLearning~\cite{tklr2021} in ITP for token reduction on 3DPW~\cite{vonMarcard2018} and Human3.6M~\cite{h36m_pami}. }
\label{tab:ITP_comparison}
\begin{center}
\resizebox{0.485\textwidth}{!}{
\setlength{\tabcolsep}{0.5mm}{
\begin{tabular}{l|c|ccccc}
\hline
\multirow{2}{*}{Method} & \multirow{2}{*}{GFLOPs} & \multicolumn{3}{c}{3DPW} & \multicolumn{2}{l}{Human3.6M} \\

&   & MPVE      &    MPJPE   & PAMPJPE     &      MPJPE     &  PAMPJPE        \\
\hline
 TokenLearner~\cite{tklr2021} &   $5.7$         &   $99.3$    & $82
 .4$       & $52.6$     & $61.2$           &  $45.4$\\
 Image Token Pruning  &   $\textbf{5.3}$     &   $\textbf{99.2}$    &    $\textbf{82.4}$   &  $\textbf{52.3}$     &  $\textbf{59.8}$        &  $\textbf{40.5}$\\
                  \hline
\end{tabular}
}
}
\end{center}
\end{table}
We compare ITP with another popular pruning strategy Tokenlearner~\cite{tklr2021} for HMR on Human3.6M and 3DPW in Table~\ref{tab:ITP_comparison}.
We use Encoder-Decoder structure~\cite{cho2022FastMETRO} with ResNet-50~\cite{he2016deep} as a backbone. The pruning rate is $20\%$.  In Table~\ref{tab:hand_mesh_recovery}, ITP saves more  computational costs while achieving the highest accuracy in terms of both human mesh recovery and joint estimation.
\subsection{NSR v.s. GCN~\cite{kipf2016semi} and MLP}
We discuss the effectiveness of NSR in GTR. We compare NSR with other implementations, including MultiLayer Perceptron and Graph Convolutional Network~\cite{kipf2016semi} following Pose2Mesh~\cite{choi2020pose2mesh}. We condition the backbone on ResNet-50~\cite{he2016deep} using Encoder-Decoder structure~\cite{cho2022FastMETRO}. In Table~\ref{supp_tab:ablation_NSR}, NSR achieves higher performance on Human3.6M and 3DPW.  This indicates that the attention mechanism in NSR provides a stronger modeling capability of vertices given the learned body features, which thus improves the quality of recovered mesh vertices.

\begin{table}[!h]
\small
\begin{center}
 \caption{Comparison of different network structures of NSR for GTR on 3DPW~\cite{vonMarcard2018} and Human3.6M~\cite{h36m_pami}. }
 \label{supp_tab:ablation_NSR}
\resizebox{0.485\textwidth}{!}{
\setlength{\tabcolsep}{0.3mm}{
\begin{tabular}{l|ccc|cc}
\hline
\multirow{2}{*}{Model} & \multicolumn{3}{c|}{3DPW} & \multicolumn{2}{c}{Human3.6M}\\
  & MPVE & MPJPE & PAMPJPE  & MPJPE & PAMPJPE \\
  \hline
  Multi-Layer Perceptron & $99.0$ & $80.4$ & $49.6$ & $57.7$ &  $38.9$  \\
Graph Convolutional Network &
 $98.8$ &  $81.5$ & $49.8$ & $58.2$ &  $38.8$  \\
Neural Shape Regressor& $\textbf{95.9}$ & $\textbf{79.2}$ & $\textbf{49.2}$ & $\textbf{57.2}$ &$\textbf{38.6}$   \\
 \hline
\end{tabular} 
  }
  }
\end{center}
 \end{table}

\subsection{Generalization on Hand Mesh Recovery}
\label{ablation:hand_mesh_recovery}
To investigate the generalizability of our framework, we conduct an experiment on monocular hand mesh recovery, which is summarized in Table~\ref{tab:hand_mesh_recovery}. Following FastMETRO~\cite{cho2022FastMETRO}, we report PAMPJPE, F-score@15mm~(F@15mm) on FreiHand~\cite{zimmermann2019freihand} together with GFLOPs$^\text{T}$ and throughput. The qualitative results are provided in Figure~\ref{fig:hand_mesh_main_result}. The hand vertex-joint interactions are visualized in Appendix~C.

\begin{table}
\small
\begin{center}
 \caption{Comparison with the SOTA methods on hand mesh recovery on FreiHAND~\cite{zimmermann2019freihand}.}
 \label{tab:hand_mesh_recovery}
\resizebox{0.485\textwidth}{!}{
\setlength{\tabcolsep}{0.25mm}{
\begin{tabular}{l|ccccc}
\hline
Method  & GFLOPs$^\text{T}$~$\downarrow$ & Throughput~$\uparrow$ & PAMPJPE~$\downarrow$  & F@15mm~$\uparrow$\\
  \hline
METRO-H64~\cite{lin2021end}  & $13.1$& $186.7$ & $6.8$ &$0.981$\\
FastMETRO-H64~\cite{cho2022FastMETRO} &$3.3$ & $228.4$ & $\textbf{6.5}$ &$\textbf{0.982}$\\
FastMETRO-H64+GTR+ITP@$20\%$  & $\textbf{1.0}$ & $\textbf{260.2}$ & $6.7$ & $0.980$\\
\hline
\end{tabular}
}
 }
\end{center}
 \end{table}
\subsection{Further Discussion}
\label{ablation:further_discussion}
\subsubsection{Analysis of Image Token Pruning}
\label{ablation:prune_vis}

\begin{figure}
\centering
\begin{overpic}[width=\linewidth]{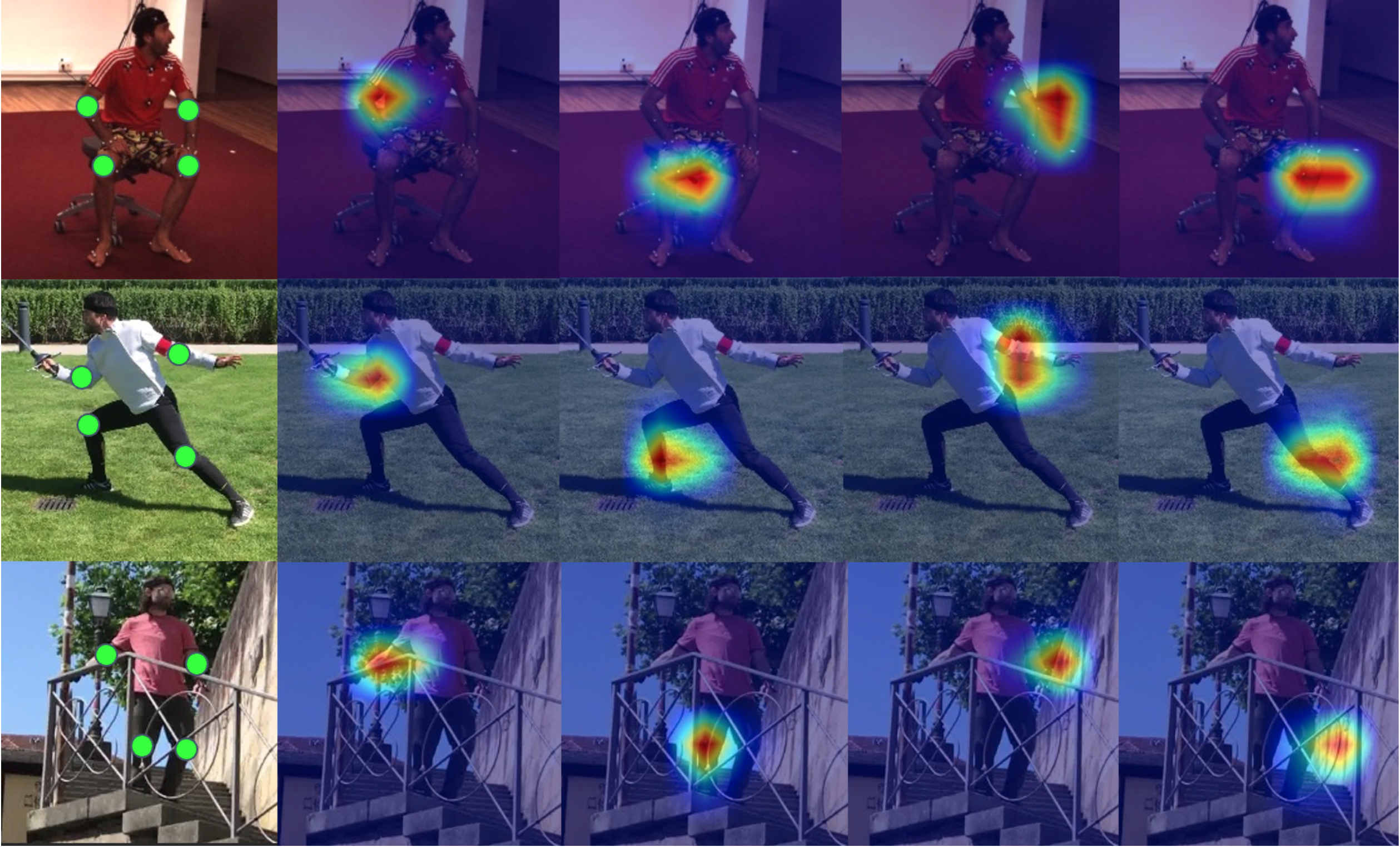}\put(6,-4){Input} 
\put(24,-4){R-Elbow}
\put(44,-4){R-Knee}
\put(64,-4){L-Elbow}
\put(84,-4){L-Knee}
\end{overpic}
\vspace{1mm}
\caption{Visualization of learned semantics by ITP. Note that the clustering results are consistent across different identities, i.e, different clusters correspond to different body joints.}
\label{fig:clustering_score}
\end{figure}
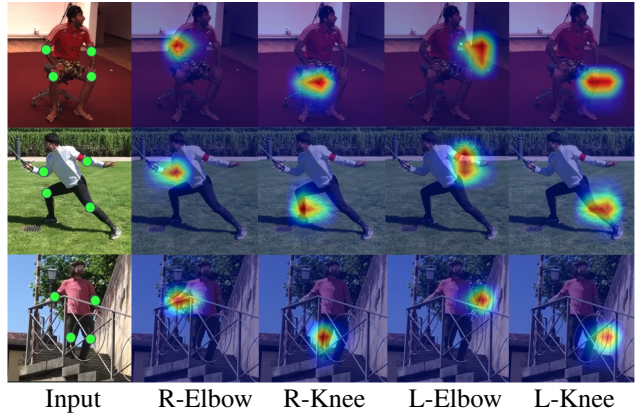

Image Token Pruning, in essence, achieves body-aware clustering results encoding discriminative body features. We visualize the heatmap of the predicted clustering scores in Figure~\ref{fig:clustering_score}, where the model is with ITP at $50\%$ pruning rate (resulting in $24$ image tokens) and GTR. The backbone is ResNet-50. As shown in Figure~\ref{fig:clustering_score}, clustering-based pruning maps original tokens to a fewer number of clusters with semantics corresponding to the body joints. Note that the semantics is consistent across different identities.

\subsubsection{ITP with Token Reduction Supervision}
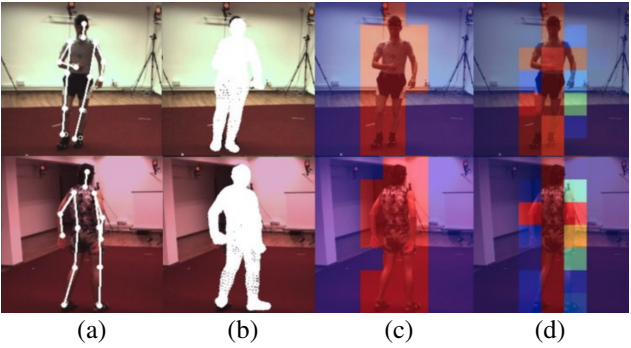
\begin{figure}
\centering
\begin{overpic}[width=\linewidth]{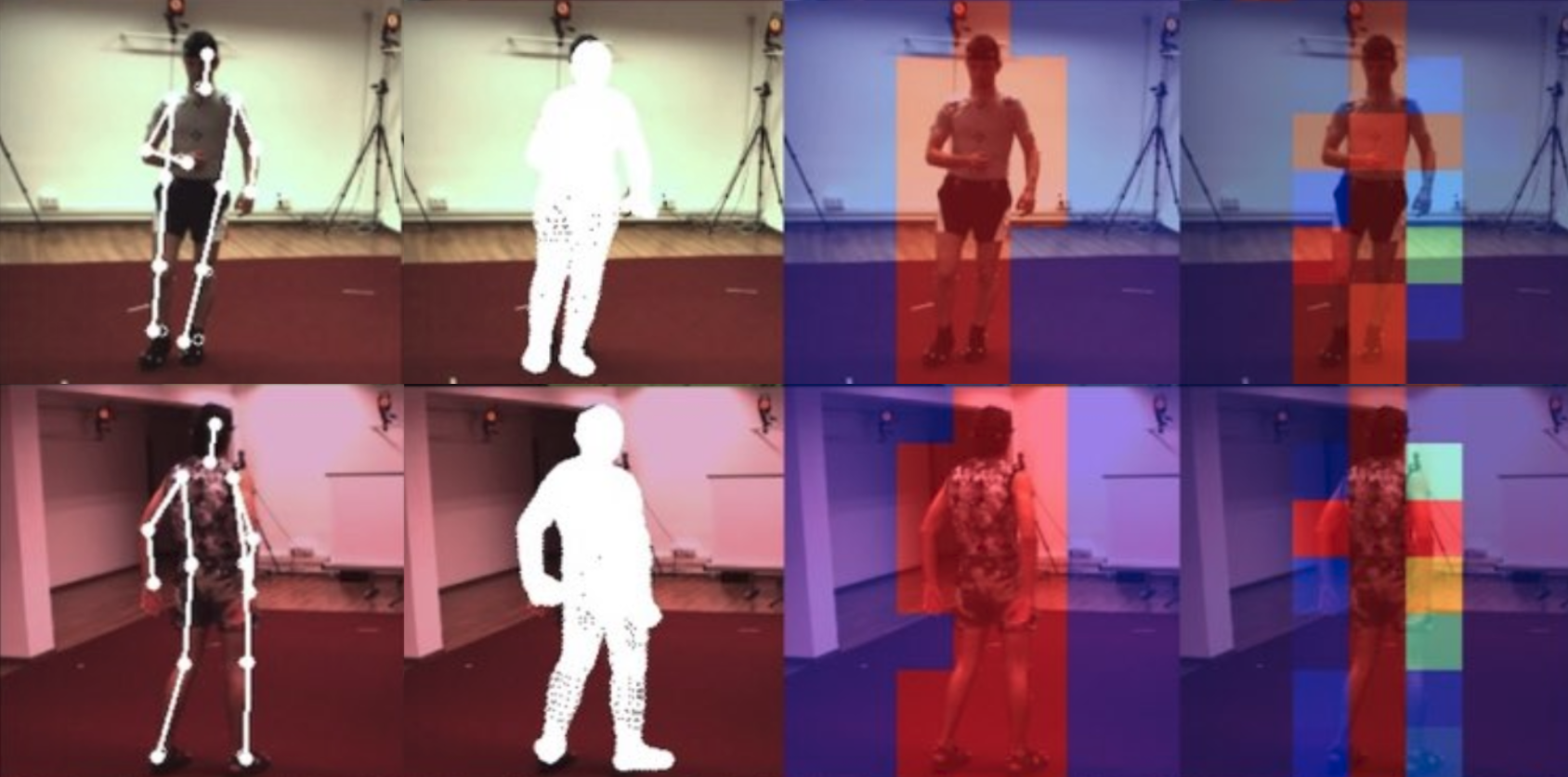}
\put(12,-4){(a)} 
\put(36,-4){(b)}
\put(61,-4){(c)}
\put(85,-4){(d)}
\end{overpic}
\vspace{1mm}
\caption{Visualization of learned semantics by Image Token Pruner. (a) projected joints. (b) projected mesh vertices. (c) mask supervision. (d) scores predicted by ITP.}
\label{fig:mask_supervision}
\end{figure}

In Figure~\ref{fig:mask_supervision}, we visualize the scores predicted by ITP in HMR. The mask supervision is generated by projecting mesh vertices to image patches using estimated camera parameters as shown in Figure~\ref{fig:mask_supervision} (b)(c). ITP effectively learns to pay attention to a human body region in the given image. Quantitatively, the projection supervision improves PAMPJPE from $41.6$ to $40.6$ on Human3.6M.

\subsubsection{Vertex-Body Feature Interactions}
\label{ablation:VJ_interaction}
\begin{figure}
\centering
\begin{overpic}[width=1.0\linewidth]{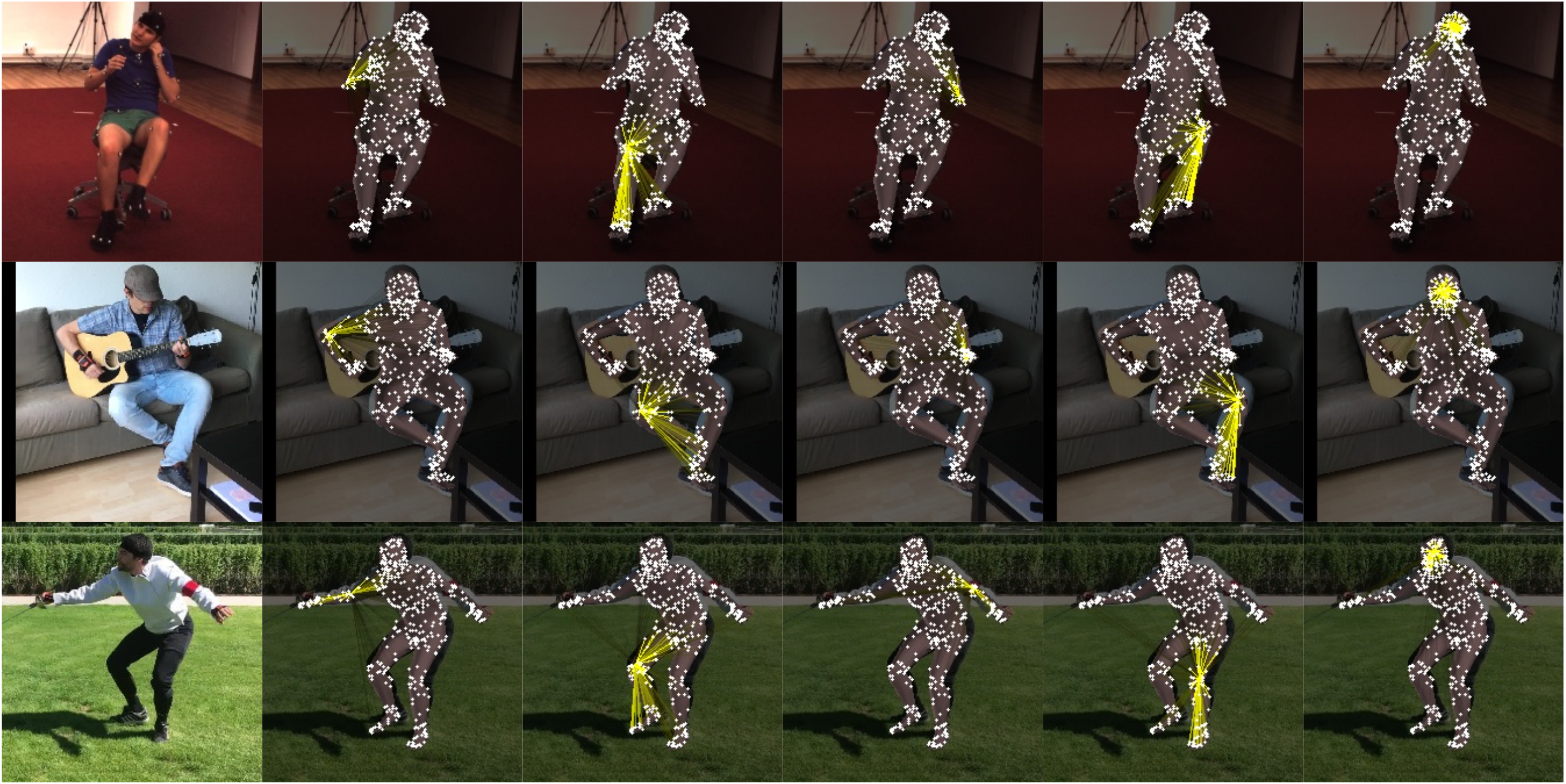} 
\put(5,-4){Input} 
\put(20,-4){R-Elbow}
\put(37,-4){R-Knee}
\put(53,-4){L-Elbow}
\put(70,-4){L-Knee}
\put(88,-4){Head}
\end{overpic}
\vspace{1mm}
\caption{Visualization of cross-attention between joint and vertices. Samples are from Human3.6M and 3DPW.}
\label{fig:V_J_interaction}
\end{figure}
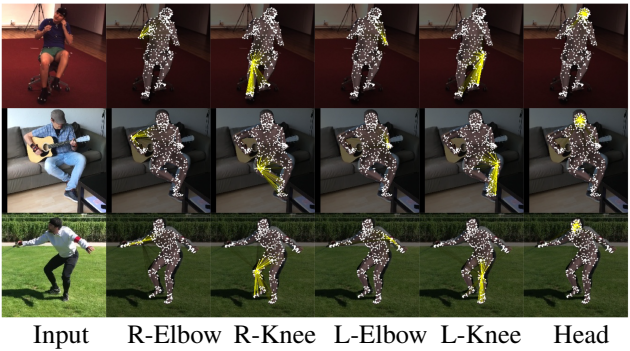

We show the interactions between vertices and joints modeled by cross attention within Neural Shape Regressor; See Figure~\ref{fig:V_J_interaction}, where the heatmap is obtained by averaging attention scores across all heads of the multi-head cross-attention between query vertices and body features. As shown in Figure~\ref{fig:V_J_interaction}, the interactions of mesh vertices and joints are at the body part level, e.g., elbow, knee, head, which are similar to the way of the blending of a human body model, i.e., SMPL~\cite{loper2015smpl} and thus validates our claims. 

\subsection{Limitations and Future Work}

When an extremely high pruning rate is applied to ITP, the accuracy of the model drops dramatically, e.g, when the token number is pruned to be one, the accuracy of the model drops dramatically ($12.1\%$) from $38.6$ to $43.3$ PAMPJPE. For GTR, since we regress joints and vertices progressively, the quality of the recovered vertices by NSR depends on the learned body features. More failure cases are in Appendix~D. 
In future work, one of the promising directions could be applying the shown enhanced efficiency of HMR from monocular images to methods exhibiting high complexity for improving the model efficiency, especially in tasks such as human-environment/object interaction that perceives environments~\cite{li2022mocapdeform, rich, prox, xia2023lightweight, Wu_2021_CVPR, hassan2021populating, shen2023learning}, as well as HMR from videos that involve temporal information~\cite{kocabas2020vibe, pavlakos2022human, lee2021uncertainty,su2020mulaycap, yuan2022glamr, wei2022capturing}. 

%% file: Main/Conclusion.tex
\section{Conclusion}
In this paper, we investigate the issues of token redundancy in the existing Transformer-based methods for both body and hand mesh recovery tasks. To tackle the problem, we introduce two effective token reduction strategies for Transformers by incorporating insights from both 3D geometry structure and 2D image features. Specifically, we recover the body shape in a hierarchical manner and cluster for image features to feed fewer but more discriminative tokens to the Transformer. Our method dramatically reduces the high-complexity interactions in the Transformer, improving the Pareto-front of accuracy and efficiency. Extensive experiments validate the proposed strategies.

\section{Acknowledgements} 
The authors would like to thank Gene Team for the fruitful discussion and the anonymous reviewers for their valuable comments and suggestions. This research is supported by Meta Reality Labs, Innovation and Technology Commission~(Ref: ITS/319/21FP) and Research Grant Council~(Ref: 17210222).

%% file: Main/Appendix.tex
\clearpage
\appendix
\renewcommand\thefigure{\Alph{section}\arabic{figure}}   
\renewcommand\thetable{\Alph{section}\arabic{table}}
This supplementary material covers: network structures and implementation details for both the Encoder-based Transformer (Sec.~\ref{supp:metro_details_enc}) and Encoder-Decoder-based Transformer (Sec.~\ref{supp:network_structure_enc_dec}); more comprehensive statistics on model efficiency (Sec.~\ref{supp:Efficiency_More}); visualization of hand vertex-joint interactions (Sec.~\ref{supp:hand_vj_interaction}); failure cases (Sec.~\ref{supp:failures}); visualization of Self-Attention within body joints and vertices (Sec.~\ref{supp:self_attention}); more qualitative comparisons with the state-of-the-art methods~(Sec.~\ref{supp:comparison}); more comparisons with existing token reduction methods~(Sec.~\ref{supp:comparison_table}) as well as discussion on pruning rate in ITP~(Sec.~\ref{supp:pruning}).

\section{Network Structure and Implementation Details}
\label{supp:network_structure_and_details}

\subsection{Transformer Encoder Structure}
\label{supp:metro_details_enc}
\paragraph{Pipeline}

\begin{figure}[!h]
\centering
\begin{overpic}[width=1.05\linewidth]{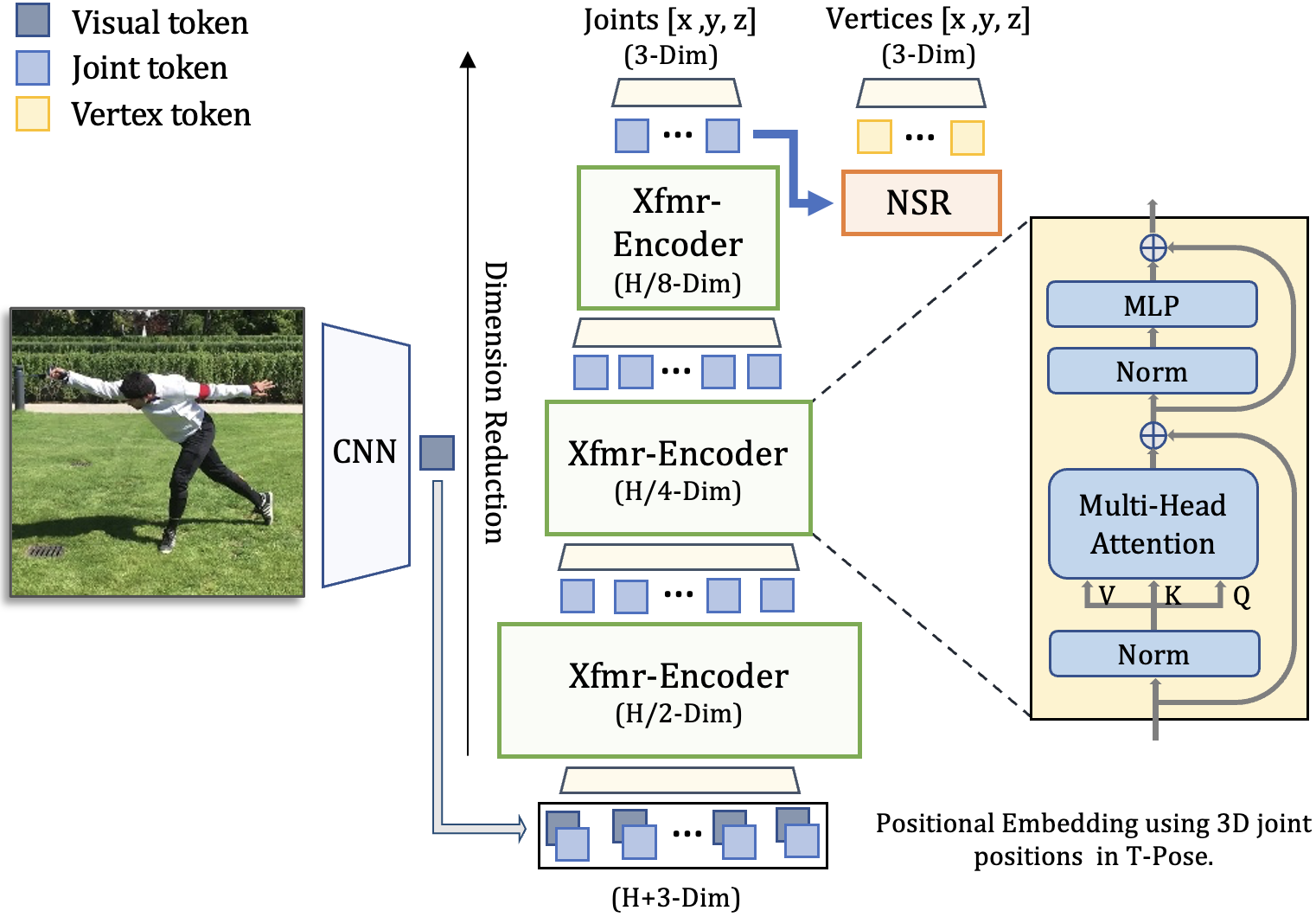}
\end{overpic}
\caption{Encoder-based Transformer~\cite{lin2021end} with  Geometry Token Reduction using Neural Shape Regressor. $H$ denotes the dimension of the feature vector.}
\label{supp_fig:enc_network}
\vspace{-6mm}
\end{figure}

We present the Geometry Token Reduction~(GTR) equipped Transformer Encoder Structure based on METRO~\cite{lin2021end} in Figure~\ref{supp_fig:enc_network}. The Transformer~(Xfmr) Encoder structure is identical to that of METRO~\cite{lin2021end}, with each block comprising a Multi-Head Attention module consisting of $4$ layers and $4$ attention heads. We employ progressive dimension reduction to decrease the hidden embedding dimensions gradually. However, note that when GTR is utilized, only joint tokens are involved in dimension reduction, and random masking over joint queries is also applied. To query the mesh vertices, NSR uses the learned joint features with $128$ feature dimensions produced by the last Xfmr Encoder block and recovers the mesh vertices. For more information on the NSR structure, refer to Sec.~\ref{supp:network_structure_enc_dec}. The CNN backbones~\cite{he2016deep,wang2020hrnet} are initialized with ImageNet-pretrained weight, and the extracted image feature size is $2048$. The positional encoding is identical to that of \cite{lin2021end}.

\subsection{Transformer Encoder-Decoder Structure}
\label{supp:network_structure_enc_dec}

An overview of the TORE-equipped Transformer Encoder-Decoder structure has been shown in Figure~1 in the main paper. To reduce dimensionality within the Transformer structure, we follow the approach of FastMETRO~\cite{cho2022FastMETRO}, where we reduce the feed-forward dimension and model dimension from $2048, 512$ to $512, 128$, respectively. The camera token and joint tokens in the Transformer Encoder and Decoder have a dimension size of $512$. The Transformer structure of FastMETRO and FastMETRO(S) has $3$ and $1$ layers, respectively. In the subsequent sections, we elaborate on the Neural Shape Regressor (NSR) and Image Token Pruner (ITP).

\begin{figure}[h!]
\centering
\begin{overpic}[width=\linewidth]{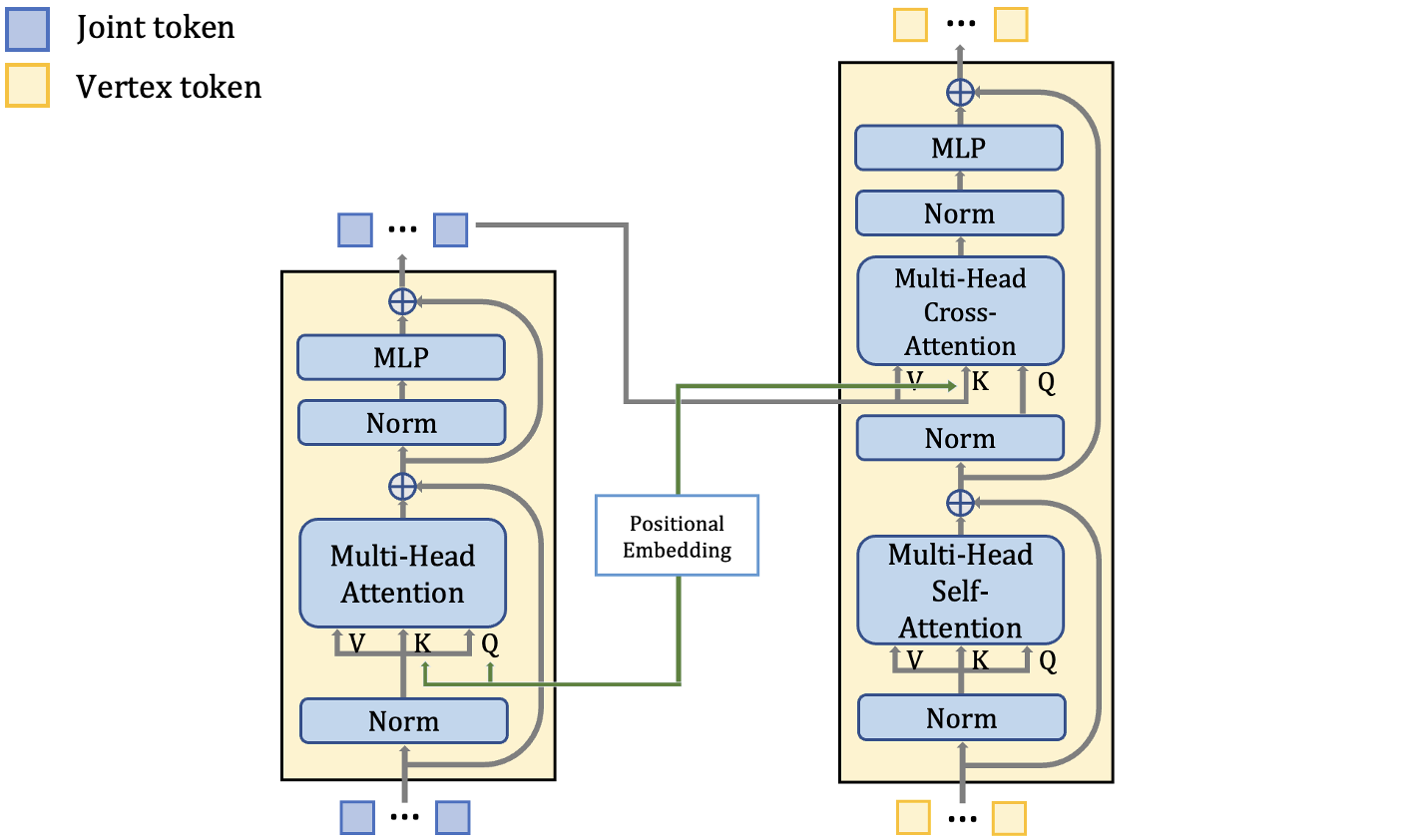}
\end{overpic}
\vspace{1mm}
\caption{Network of Neural Shape Regressor.}
\label{supp_fig:enc_dec_network}
\end{figure}

\paragraph{Neural Shape Regressor}
Herein, we present the Neural Shape Regressor (NSR) structure, which is implemented using a Transformer Encoder-Decoder as illustrated in Figure~\ref{supp_fig:enc_dec_network}. Initially, the joint features $F_J = \{f^j_1, f^j_2,..., f^j_J\}$, where $f^j_i \in \mathbb{R}^{128 \times 1}$ and $J=14$, are processed by a Multi-Head Self-Attention module to improve their representation. Cross-Attention is then used to learn the interaction between the vertex tokens $T_\mathcal{V} = \{t^v_1, t^v_2,..., t^v_V\}$, where $f^j_i \in \mathbb{R}^{128 \times 1}$ and $V=431$, and the learned joint features. Prior to Cross-Attention, Self-Attention is applied among the vertex query tokens, and non-adjacent vertices are masked out to enhance efficiency, as suggested by~\cite{cho2022FastMETRO}. The NSR has a feed-forward dimension and model dimension of $512$ and $128$, respectively, and employs fixed sinusoidal positional encoding~\cite{carion2020end}. The learned non-local interactions among joints and vertices can be visualized in Sec.~\ref{supp_fig:self_attn_j} and Sec.~\ref{supp_fig:self_attn_v}, respectively.

\paragraph{Image Token Pruner}

Given an input monocular image, the feature is extracted using CNN backbones~\cite{he2016deep,wang2020hrnet,tan2019efficientnet}, resulting in a feature map $F_\mathcal{I} \in \mathbb{R}^{H\times W \times C}$, where $H=7$, $W=7$, and $C=2048$. After reducing the dimensionality of $F_\mathcal{I}$ from $C$ to $C'=512$, we obtain a dimension-reduced feature map $F'_\mathcal{I} \in \mathbb{R}^{H \times W \times C'}$. Subsequently, we flatten $F'_\mathcal{I}$ to $F'_\mathcal{I} \in \mathbb{R}^{HW \times C'}$ to create $HW$ tokens, which are passed to our ITP module to reduce computational costs in the transformer model.

\paragraph{Training Details}
TORE-equipped FastMETRO~\cite{cho2022FastMETRO} utilizes the same loss terms as described in Sec. \ref{supp:metro_details_enc}. The AdamW optimizer is applied with a learning rate and weight decay of $10^{-4}$. Gradient clipping is implemented with a maximal gradient norm value of $0.3$. As with METRO~\cite{lin2021end}, the weights of the CNN backbones~\cite{he2016deep,wang2020hrnet} are initialized using ImageNet-pretrained weights. The low and high-resolution meshes used for hand mesh recovery consist of $195$ and $778$ vertices, respectively.

\section{More Statistics on Model Efficiency}
\label{supp:Efficiency_More}
\begin{figure*}
\centering
\begin{overpic}[width=0.98\linewidth]{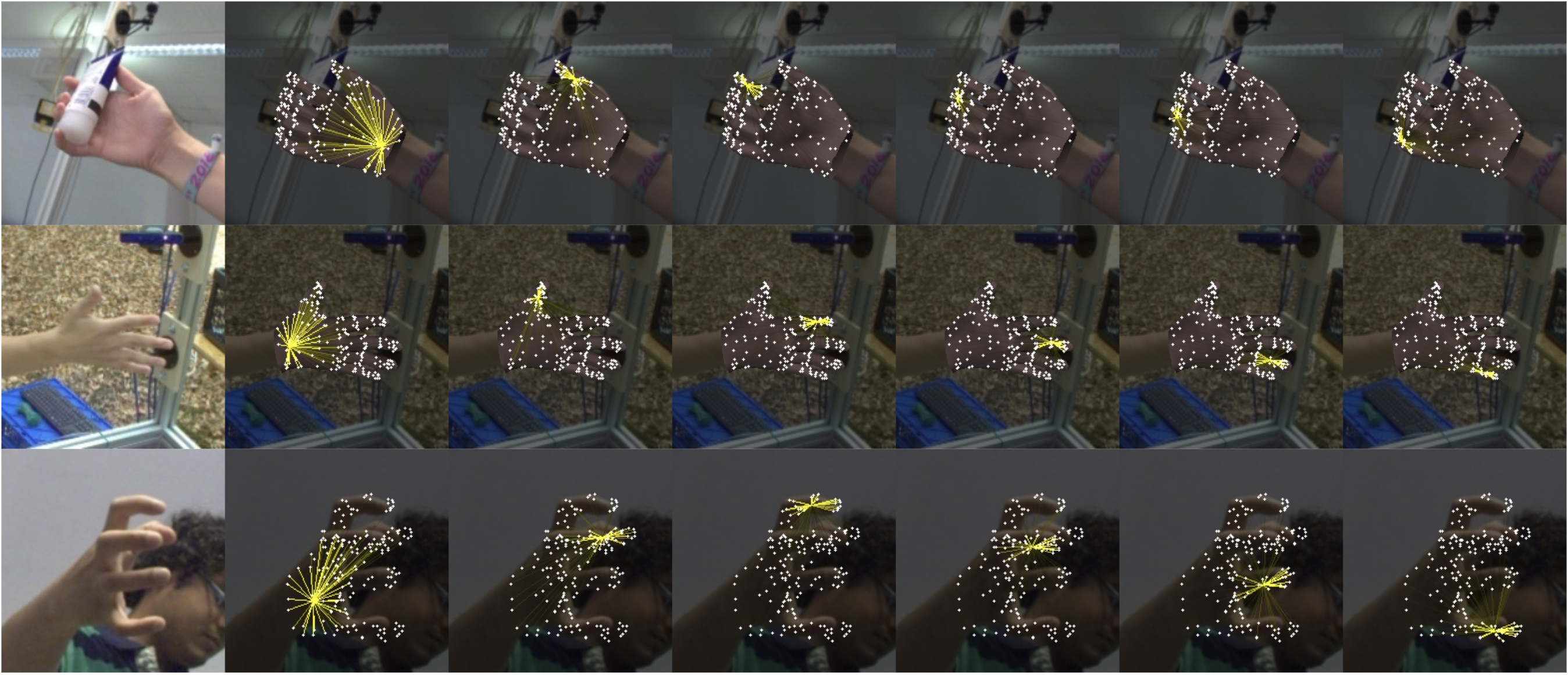}
\put(5,-2){Input} 
\put(19.5,-2){Wrist}
\put(33,-2){Thumb}
\put(48,-2){Index}
\put(61.5,-2){Middle}
\put(77,-2){Ring}
\put(91,-2){Pinky}
\end{overpic}
\vspace{3mm}
\caption{Cross-Attention among hand joints and hand vertices.}
\label{supp_fig:hand_attn}
\vspace{-5mm}
\end{figure*}
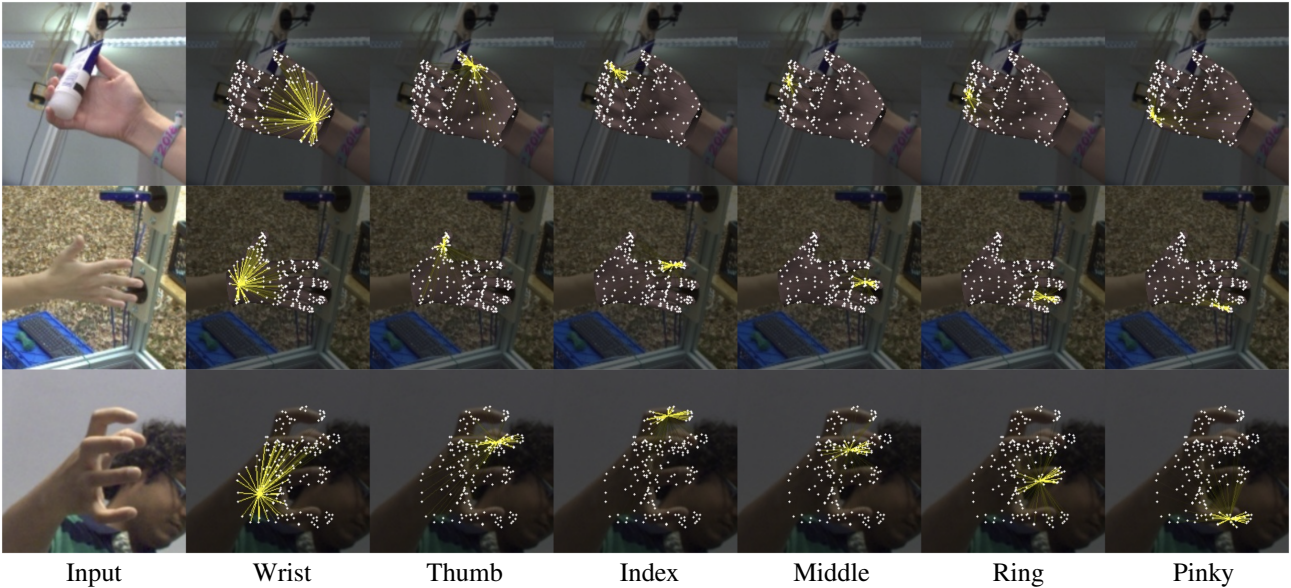

\subsection{Model Efficiency at Full Memory Usage}
In order to thoroughly examine the capabilities of our models, we perform a throughput analysis using the maximum batch size possible. Our approach involves attempting to fit the largest possible batches into the VRAM of a GPU, specifically an RTX3090 card with $24$G VRAM, and conducting a throughput analysis based on this setting. This enables us to uncover the full potential of the models on a consumer-grade graphics card.

\begin{table}[h!]
\small
\caption{Comparison for throughput on maximum batch size for monocular 3D human mesh recovery on Human3.6M~\cite{h36m_pami}. We test with ResNet-50~\cite{he2016deep} and HRNet-W64~\cite{wang2020hrnet} as backbones.}
 \label{supp_tab:max_bs_inference}
\begin{center}
 \resizebox{0.485\textwidth}{!}{
\setlength{\tabcolsep}{1mm}{
\begin{tabular}{l|c|c|c}
\hline
{Method} & {Max BS} &  {GFLOPs~ $\downarrow$} &  {Throughput~ $\uparrow$} \\
\hline
METRO-H64 & $16$ & $56.5$ &  $141$  \\
METRO-H64+GTR & $32$ & $30.3$ &  $246.6$  \\
FastMETRO-H64 & $800$ & $35.7$ & $249.7$  \\ 
FastMETRO-H64+GTR+ITP@$20\%$ & $1248$ & $\textbf{30.2}$ & $\textbf{302.3}$  \\
\hdashline
METRO-R50 & $24$ & $31.6$ &  $247$ \\
METRO-R50+GTR & $80$ & $5.4$ &  $982.5$  \\
FastMETRO-R50 & $1024$ & $10.9$ & $634.2$ \\ 
FastMETRO-R50+GTR+ITP@$20\%$ & $1120$ & $\textbf{5.3}$ & $\textbf{1086.4}$  \\ 
 \hline
\end{tabular} 
  }}
\end{center}
\vspace{-2mm}
 \end{table} 

As shown in Table~\ref{supp_tab:max_bs_inference} demonstrates that using TORE results in enhanced model effectiveness and increased inference throughput. Our approach enables larger batch sizes, higher computational throughput, and lower GFLOPs in computation when the GPU capabilities are maximized, owing to fewer tokens. This renders our approach more practically useful than the prior approach~\cite{lin2021end,cho2022FastMETRO}.

\subsection{Training Memory Cost Comparison}
\label{supp_sec:memory_cost}

We conducted experiments on the training cost per GPU VRAM to demonstrate our superiority in training resource efficiency. As presented in Table~\ref{supp_tab:comparison_gpu_memory_enc} and Table~\ref{supp_tab:comparison_gpu_memory_enc_dec}, the models that integrate our proposed TOken REduction (TORE) methods display significantly reduced GPU VRAM consumption. Specifically, with TORE, the Transformer Encoder structures~\cite{lin2021end} with ResNet-50 (R50)~\cite{he2016deep} and HRNet-W64 (H64)~\cite{wang2020hrnet} as backbones demonstrate memory savings of $58.3\%$ and $44.2\%$, respectively. Similarly, the Transformer Encoder-Decoder structures~\cite{cho2022FastMETRO} with TORE using ResNet-50 and HRNet-W64 as backbones exhibit reduced memory costs of $27.4\%$ and $17.9\%$, respectively.

\begin{table}[h!]
\small
 \caption{Comparison on Training GPU VRAM Cost of the Transformer Encoder structure~\cite{lin2021end}. The model is trained on $8$ GPU cards with a batch size to be $32$.}
 \label{supp_tab:comparison_gpu_memory_enc}
\begin{center}
\resizebox{0.4\textwidth}{!}{
\setlength{\tabcolsep}{6mm}{
\begin{tabular}{l|l}
\hline
  Model   & GPU Memory Cost\\
  \hline
METRO-H64 &  $32.8$GB \\
METRO-H64+GTR& $18.3$GB~$\textbf{(-44.2\%)}$\\
\hdashline
METRO-R50 &  $24.7$GB \\
METRO-R50+GTR&  $10.3$GB~$\textbf{(-58.3\%)}$\\

 \hline
\end{tabular}}
}
\end{center}
\end{table}

\begin{table}[h!]
\small
\caption{Comparison on Training GPU VRAM Cost of the Transformer Encoder-Decoder structure~\cite{cho2022FastMETRO}. The model is trained on $4$ GPU cards with a batch size to be $16$.}
 \label{supp_tab:comparison_gpu_memory_enc_dec}
\begin{center}
\resizebox{0.4\textwidth}{!}{
\setlength{\tabcolsep}{1mm}{
\begin{tabular}{l|l}
\hline
  Model   & GPU Memory Cost $\downarrow$\\
  \hline
FastMETRO-H64 &  $13.4$GB \\
FastMETRO-H64+GTR+ITP@$20\%$& $11.0$GB $\textbf{(-17.9\%)}$ \\
\hdashline
FastMETRO-R50 & $8.4$GB\\
FastMETRO-R50+GTR+ITP@$20\%$&  $6.1$GB $\textbf{(-27.4\%)}$\\
 \hline

\end{tabular}
}
}
\end{center}
\end{table}

\subsection{Run-time Performance}
\vspace{-1.5mm}

The run-time performance statistics of our models are presented in Table~\ref{tab:run-time}, under the same hardware configuration as Sec. 4.3 in the Main paper. 

Notably, we evaluated system performance using established metrics such as throughput and GFLOPs, as utilized in PPT~\cite{ma2022ppt}, DynamicViT~\cite{rao2021dynamicvit}, TokenLearner~\cite{tklr2021}, Evit~\cite{liang2022not}, and CrossVit~\cite{chen2021crossvit}, among others. For system performance investigation, Unlike the FPS, which considers the processing of a single instance, throughput is typically used since it measures the maximum number of input instances that the network can process in a given time unit, evaluating the parallel processing of multiple instances~\cite{ma2022ppt}.

We further discuss the following instance: 

\noindent
\textbf{Top-Down Human Mesh Recovery} During the inference process of the top-down approach in multi-person HMR, an object detector locates multiple human instances in a given input image, which are typically cropped, resized, and grouped into a minibatch for faster inference. The resulting minibatch is then fed into the pose detector. In this common case, we consider throughput a more appropriate metric for evaluating the performance of top-down HMR tasks.

\noindent
\textbf{Multi-Camera System}
In practical use of a multi-camera system, e.g., surveillance, sports analysis, and crowd management, multiple camera feeds are sent to a centralized server for analysis. To efficiently process the aggregated images from multiple cameras, high throughput is required. A high throughput ensures that the system can simultaneously process the camera feeds without any lag or delay in these scenarios. For those offline applications where real-time performance is not highly demanded, a high throughput system also saves time and cost.

\begin{figure*}
\centering
\begin{overpic}[width=\linewidth]{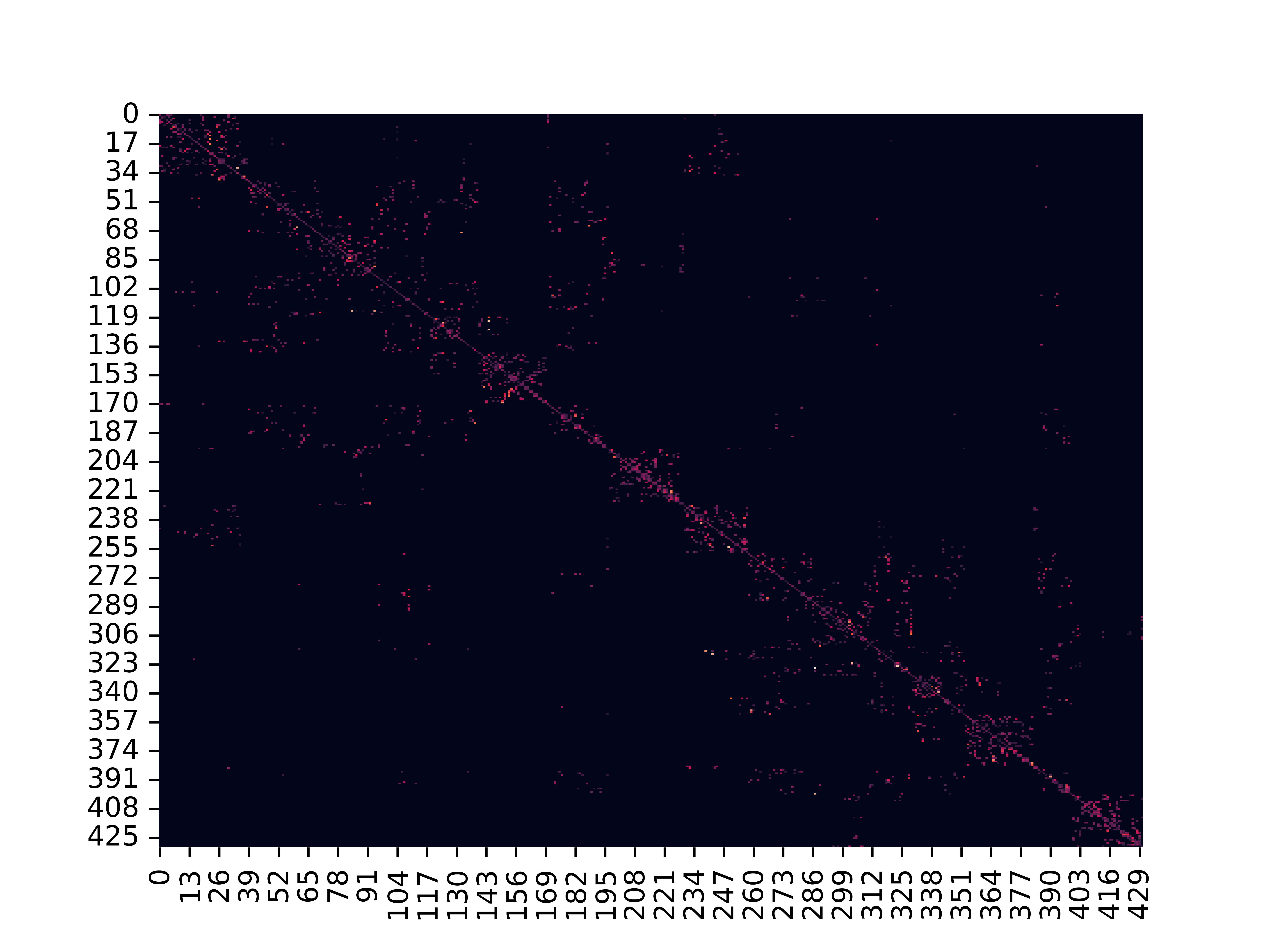}
\end{overpic}
\caption{Visualization of Self-Attention within the body vertices.}
\label{supp_fig:self_attn_v}
\end{figure*}

\section{Vertex-Joint Interactions in Hand Mesh Recovery}
\label{supp:hand_vj_interaction}

Figure~\ref{supp_fig:hand_attn} demonstrates the interactions modeled by the Neural Shape Regressor (NSR) on Hand Mesh Recovery. To obtain attention scores, we average the scores across all heads of the multi-head cross-attention between query vertices and joint features. The interactions between mesh vertices and joints in the hand model exhibit a shape-blending style similar to MANO~\cite{MANO:SIGGRAPHASIA:2017}. This observation aligns with the human body model and validates the effectiveness of our proposed methods. We used the FastMETRO-H64+GTR+ITP@$20\%$ model for visualization.

\begin{table}[!h]
\small
 \caption{Running time Performance (FPS).}
 \label{tab:run-time}
 \vspace{3mm}
 
\begin{center}
\resizebox{0.4\textwidth}{!}{
 \setlength{\tabcolsep}{5mm}{
\begin{tabular}{l|c}
\hline
{Method} & FPS \\
\hline
METRO-H64+GTR & $22$\\
FastMETRO-H64+GTR+ITP@$80\%$ & $26$ \\
\hdashline
METRO-R50+GTR & $53$\\
FastMETRO-R50+GTR+ITP@$80\%$ & $61$ \\
 \hline
\end{tabular}
}
}
\end{center}
\end{table}

\begin{figure*}
\centering
\begin{overpic}[width=0.45\linewidth]{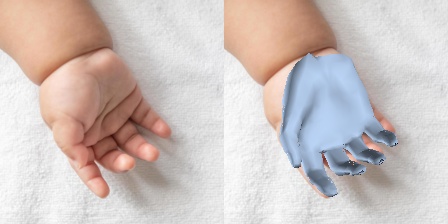}
\put(48.5,-5){(a)} 
\end{overpic}
\begin{overpic}[width=0.45\linewidth]{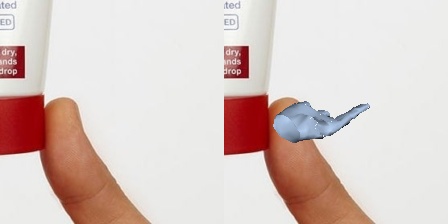}
\put(48.5,-5){(b)} 
\end{overpic}
\vspace{7mm}
\caption{Failure cases when the inputs are outside of the
training data distribution.}
\label{supp_fig:failure_cases_hand}
\end{figure*}
\section{Failure Cases}
\label{supp:failures}

\begin{figure}
\centering
\begin{overpic}[width=0.98\linewidth]{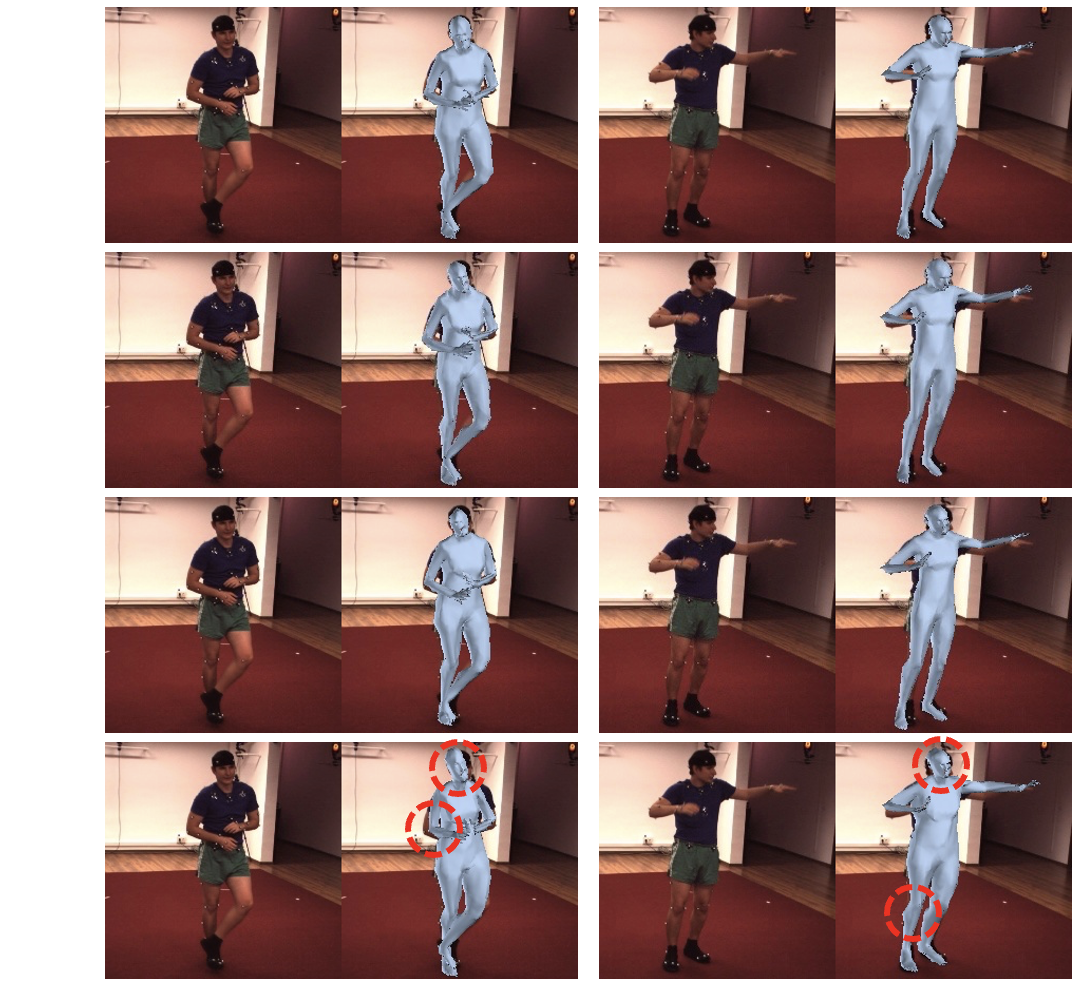}
\put(0,80){$0\%$} 
\put(0,56){$5\%$} 
\put(0,34){$7.5\%$}
\put(0,10){$10\%$} 
\put(25,-3){Case 1} 
\put(72,-3){Case 2}
\end{overpic}
\vspace{3mm}
\caption{Human mesh recovery results at different noise levels to NSR during GTR. We showcase human mesh recovery at four noise levels: $0\%$, $5\%$, $7.5\%$ and $10\%$.}
\label{supp_fig:failure_cases_gtr}
\vspace{-5mm}
\end{figure}

Since Geometry Token Reduction  recovers human mesh hierarchically, the quality of the recovered vertices by NSR depends on the learned body features. We conducted two experiments by adding Gaussian noise $\epsilon \in \mathbb{R}^K$ (at $5\%, 7.5\%$, $10\%$) to NSR to the body features for mesh vertex regression. $K$ is the dimension of input features. As shown in Figure~\ref{supp_fig:failure_cases_gtr}, when the body features are unreliable, e.g., $10\%$ noise level, the performance of human mesh recovery by GTR drops.

Additionally, when the input image is outside of the training data distribution, such as a photo of an infant's hand, the recovered mesh's quality may diminish, as shown in Figure~\ref{supp_fig:failure_cases_hand} (a). Moreover, when the model encounters extremely challenging partial observation, such as an image capturing only one thumb, it cannot recover the hand mesh accurately; see Figure~\ref{supp_fig:failure_cases_hand} (b). The model used for this study was FastMETRO-H64+GTR+ITP@$20\%$.

\section{Self-Attention within Joints and Vertices}
\label{supp:self_attention}
\begin{figure}[!h]
\centering
\begin{overpic}[width=\linewidth]{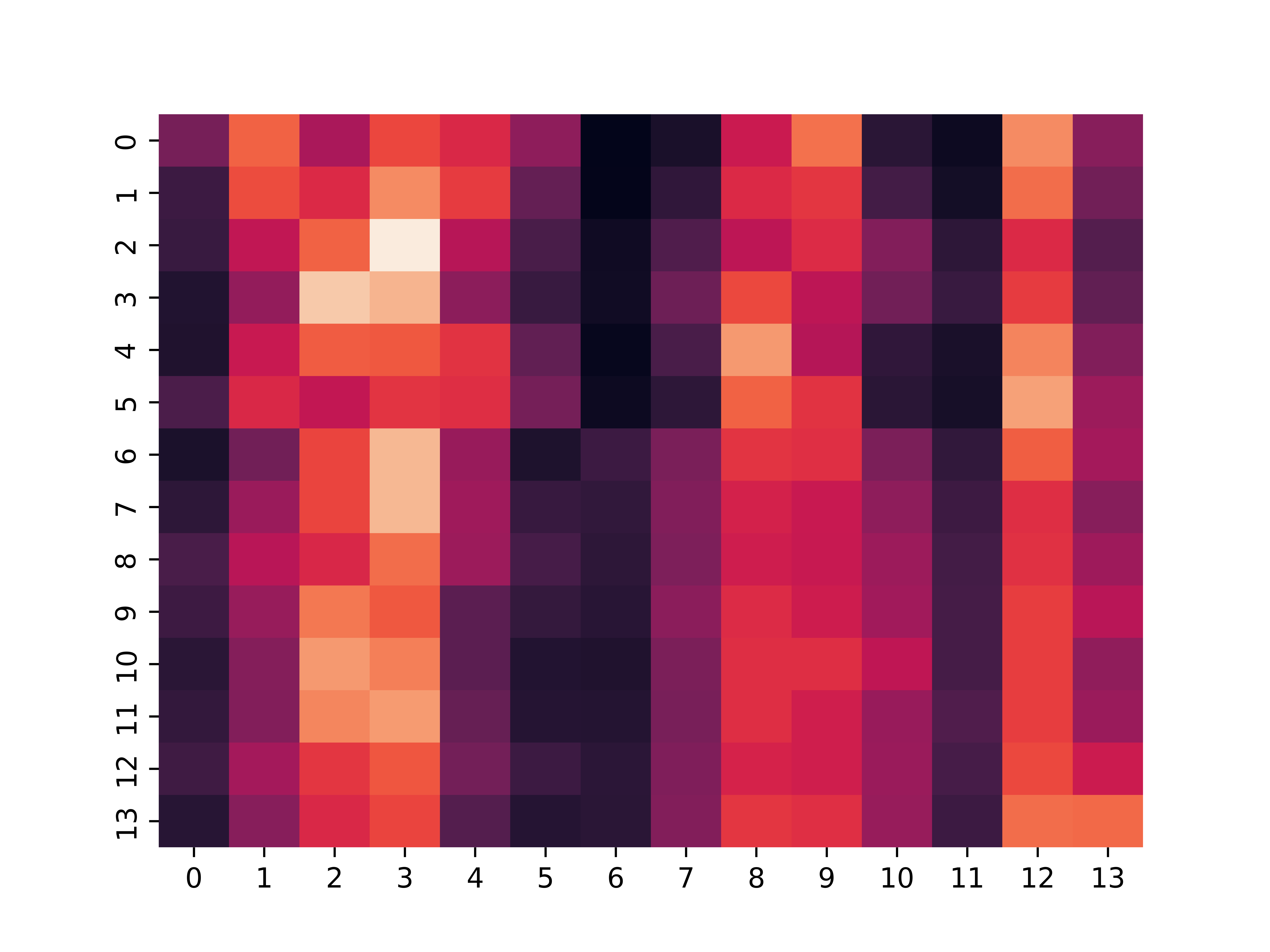}
\end{overpic}
\caption{Visualization of Self-Attention within the body joints}
\label{supp_fig:self_attn_j}
\end{figure}

In this section, we investigate the interactions between joints and vertices within the Self-Attention module. Specifically, joint interactions are analyzed by averaging attention scores from all Multi-Head Self-Attention modules within the Transformer Encoder, while vertex interactions are assessed by averaging scores from all heads of Multi-Head Self-Attention within the Transformer Decoder. The resulting Self-Attention visualizations for joints and vertices are presented in Figure~\ref{supp_fig:self_attn_j} and Figure~\ref{supp_fig:self_attn_v}, respectively. As depicted in Figure~\ref{supp_fig:self_attn_j}, Self-Attention effectively models non-local interactions among joints, thus improving model robustness to partial observation and self-occlusion in challenging monocular observations. Similarly, Figure~\ref{supp_fig:self_attn_v} shows that non-local interactions among vertices enhance mesh recovery performance when learned joint features are employed. Visualization was performed using FastMETRO-H64+GTR+ITP@$20\%$.

\section{More Qualitative Comparisons}
\label{supp:comparison}
\begin{figure*}
\centering
\begin{overpic}[width=0.86\linewidth]{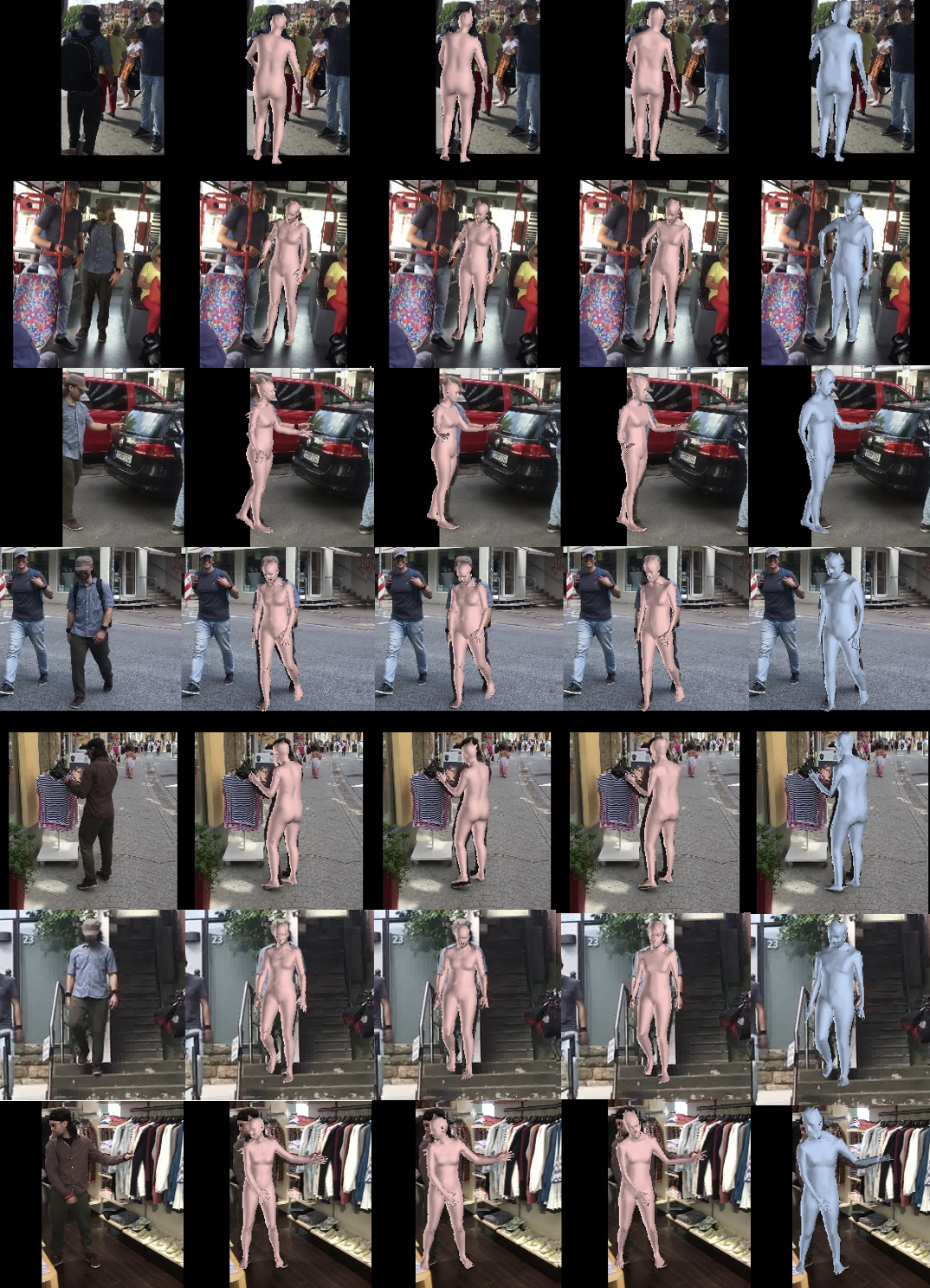}
\put(5,-3){Input} 
\put(17,-3){METRO~\cite{lin2021end}}
\put(29,-3){MeshGraphormer~\cite{lin2021mesh}}
\put(46,-3){FastMETRO~\cite{cho2022FastMETRO}}
\put(63,-3){Ours}
\end{overpic}
\vspace{10mm}
\caption{Qualitative Comparison with existing Transformer-based methods~\cite{lin2021end,lin2021mesh,cho2022FastMETRO} on 3DPW~\cite{vonMarcard2018}.}
\label{supp_fig:3dpw}
\end{figure*}

\begin{figure*}
\centering
\begin{overpic}[width=0.86\linewidth]{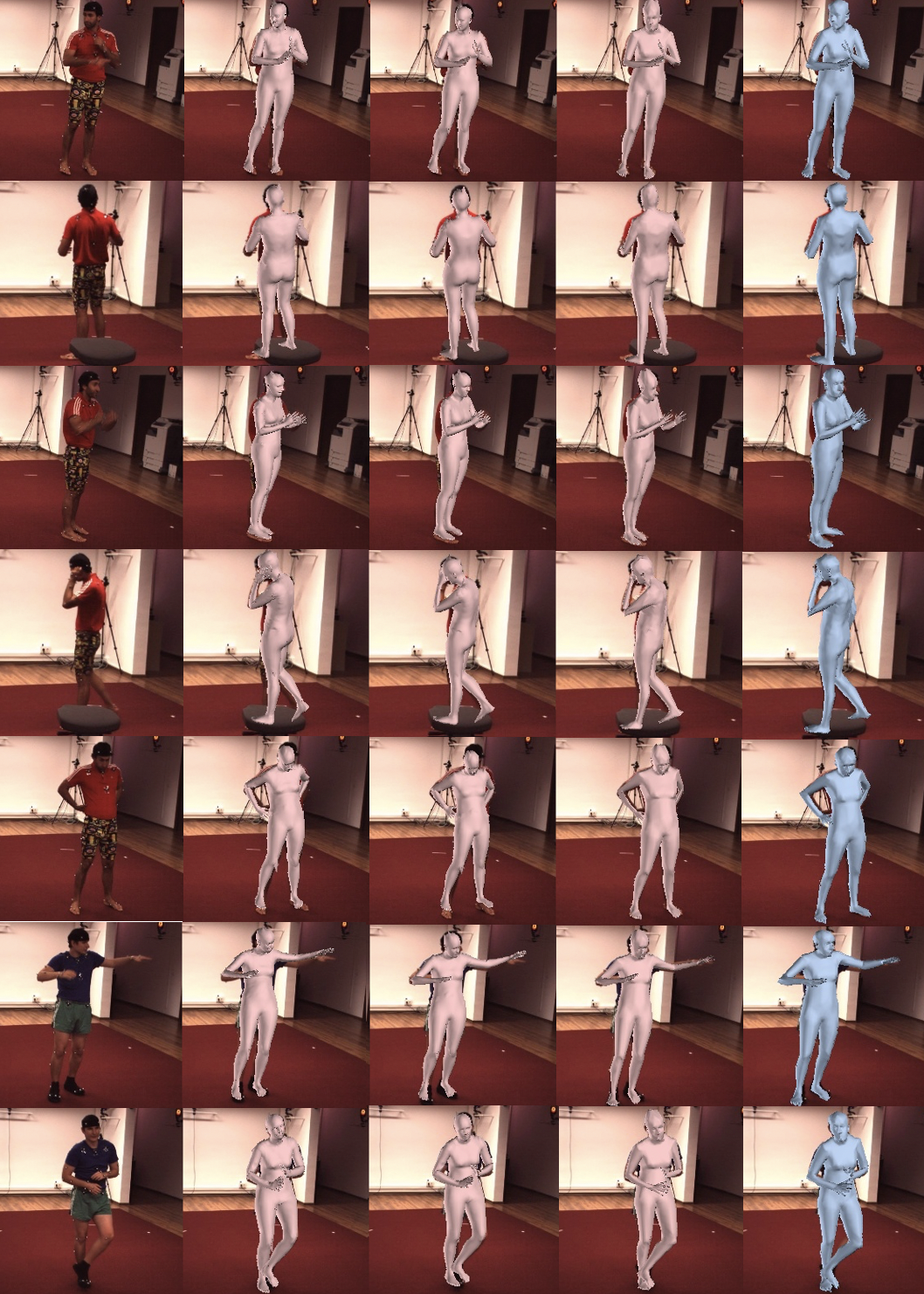}
\put(5,-3){Input} 
\put(17,-3){METRO~\cite{lin2021end}}
\put(28,-3){MeshGraphormer~\cite{lin2021mesh}}
\put(45,-3){FastMETRO~\cite{cho2022FastMETRO}}
\put(62,-3){Ours}
\end{overpic}
\vspace{10mm}
\caption{Qualitative Comparison with existing Transformer-based methods~\cite{lin2021end,lin2021mesh,cho2022FastMETRO} on Human3.6M~\cite{h36m_pami}.}
\label{supp_fig:h36m}
\end{figure*}

The introduction of TORE greatly saves the computational cost, i.e., GFLOPs and improves the throughput while enabling the model to produce competitive or even better mesh recovery from monocular images. We further conducted qualitative comparisons with existing methods such as ~\cite{lin2021end, lin2021mesh, cho2022FastMETRO} for human mesh recovery on 3DPW~\cite{vonMarcard2018} and Human3.6M~\cite{h36m_pami}. Our results are summarized in Figure~\ref{supp_fig:3dpw} and Figure~\ref{supp_fig:h36m}, respectively. All methods utilized HRNet-W64~\cite{wang2020hrnet} as the CNN backbone, and our model setting is FastMETRO-H64+GTR+ITP@$20\%$.

We also conduct qualitative comparisons with existing methods~\cite{hendrycks2016gaussian, lin2021mesh, cho2022FastMETRO} for human mesh recovery on 3DPW~\cite{vonMarcard2018} and Human3.6M~\cite{h36m_pami}, which are summarized in Figure~\ref{supp_fig:3dpw} and Figure~\ref{supp_fig:h36m}, respectively. All methods use the HRNet-W64~\cite{wang2020hrnet} as a CNN backbone, and our model setting is FastMETRO-H64+GTR+ITP@$20\%$.

\section{More Comparisons with Existing Methods}
\label{supp:comparison_table}
\subsection{Comparison with TCFormer~\cite{zeng2022not}.}
We conduct a comparison between TORE and another token clustering method for HMR. Compared with TCFormer~\cite{zeng2022not}, we have \textit{1) Different architectures}. TCFormer is a much more complicated multi-stage method for token clustering, while ours only requires a single pass; \textit{2) Different body representation}. There is no consideration of body representation in TCFormer while we propose NSR for GTR to reduce redundancy; \textit{3) Different performance.} Compared with TCFormer with the same setting in Tab~\ref{rebuttal_tab:comp_tcformer}, where our method surpasses TCFormer on both two datasets. 

\begin{table}
\small
 \caption{TCFormer~\cite{zeng2022not} v.s. TORE (Ours) for HMR on Human3.6M.}
 \label{rebuttal_tab:comp_tcformer}
\begin{center}
\resizebox{0.485\textwidth}{!}{
\setlength{\tabcolsep}{1.1mm}{
\begin{tabular}{l|c|cc|cc}
\hline
\multirow{2}{*}{Method} & \multirow{2}{*}{Throughput$\uparrow$} & \multicolumn{2}{c|}{3DPW} & \multicolumn{2}{c}{Human3.6M}\\
& & MPJPE$\downarrow$ & PAMPJPE$\downarrow$  & MPJPE$\downarrow$ & PAMPJPE$\downarrow$ \\
  \hline
TCFormer~\cite{zeng2022not} & $230.9$ &$80.6$ & $49.3$ & $62.9$ & $42.8$ \\
METRO+TORE (Ours)& $ 210.1$& $75.5$ & $46.6$ & $\mathbf{57.6}$ & $37.1$\\
FastMETRO+TORE (Ours) & $ 249.2$ & $\mathbf{72.3}$	&$\mathbf{44.4}$ & $ 59.6$ & $\mathbf{36.4}$ \\
 \hline
\end{tabular} 
  }
  }
\end{center}
 \end{table} 
\subsection{Comparison with PPT~\cite{ma2022ppt}}
We compared with PPT~\cite{ma2022ppt} that prunes tokens by locating human visual tokens according to attention score; see Tab~\ref{rebuttal_tab:comparison_ppt} where ours is more competitive.

\begin{table}[H]
\small
 \caption{PPT~\cite{ma2022ppt} v.s. TORE (Ours). We test with FastMETRO-Eb0 on Human3.6M.}
  \label{rebuttal_tab:comparison_ppt}
\begin{center}
\resizebox{0.485\textwidth}{!}{
\setlength{\tabcolsep}{2mm}{
\begin{tabular}{l|c|c|c|c}
\hline
Method & GFLOPs$\downarrow$ & Throughput$\uparrow$	& MPJPE$\downarrow$ & PAMPJPE$\downarrow$ \\
\hline
PPT 	& $1.6$& $862.1$ & $68.4$ & $46.2$\\
Ours & $1.6$& $\mathbf{870.4}$ & $\textbf{63.2}$&$\mathbf{43.9}$\\
 \hline
\end{tabular} 
}
}
\end{center}
\end{table}

\section{Influence of Pruning Rate in ITP}
\label{supp:pruning}
The influence of different pruning rates is shown in Tab~\ref{rebuttal_tab:pruning}. In this paper, we empirically set the pruning rate to $20\%$ based on extensive experiments. 

\begin{table}[H]
\small
\caption{Influence of pruning rates. We test with FastMETRO-Eb0 on Human3.6M.}
 \label{rebuttal_tab:pruning}
\begin{center}
\resizebox{0.485\textwidth}{!}{
\setlength{\tabcolsep}{5mm}{
\begin{tabular}{l|cc|c}
\hline
Pruning rate & PAMPJPE $\downarrow$ & MPJPE $\downarrow$ & GFLOPS $\downarrow$ \\
\hline
No Pruning & $45.8$ & $69.2$ & $7.1$ \\
$0.2$ & $\mathbf{43.9}$ & $\mathbf{63.2}$ & $1.6$ \\
$0.5$ & $\mathbf{43.9}$ & $64.2$ & $1.4$ \\
$0.75$ & $44.7$ & $65.3$ & $\mathbf{1.2}$ \\
 \hline
\end{tabular} 
  }}
\end{center}
\end{table}

\clearpage

\clearpage